# Machine Translation with Large Language Models: Prompt Engineering for Persian, English, and Russian Directions

NOOSHIN POURKAMALI, University of Isfahan, Iran
SHLER EBRAHIM SHARIFI, University of Isfahan, Iran

Generative large language models (LLMs) have demonstrated phenomenal proficiency across several NLP tasks such as machine translation, question answering, text summarization, and natural language understanding.
To further improve the performance of LLMs on MT use cases, we investigated 2 different popular prompting methods and a combination of them a cross language combinations of Persian, English, and Russian, where we used n-shot feeding and tailored prompting frameworks. Our results indicate that LLMs like PaLM, which are trained with multilingual data can generate human-like MT outputs, can better fine-tune desired translation nuances, according to the style guidelines and other linguistics considerations, and generally are better at processing and applying prompts; Depending on the Language model, MT task on the hand, source and target language, we point out some considerations for using these models, when adopting prompting frameworks, and using n-shot in-context learning.

Additionally, we discovered some errors and limitations of popular LLMs as an MT tool and categorized these errors into a number of linguistics metrics as a typology. Our findings aim to provide a preliminary experience for appropriate usage of LLMs while offering some methods for designing prompts for in-context learning. We anticipate that this report will shed new light on advancing the field of machine translation with LLMs by enhancing both the accuracy and reliability of metrics.

## 1 INTRODUCTION

Large language models (LLMs), especially Generative Pre-trained Transformer (GPT) models [42], [4] have shown significant positive results in various natural language processing (NLP) tasks. An example, OpenAI with ChatGPT based on InstructGPT [36], has become the spotlight for NLP community researchers in the field of machine translation as a game changing tool, as well as several other NLP tasks [13], [29], [8], including question answering, content generation, evaluation, and so on [41], [66] [57], [30], [32]. All these capabilities have opened up new possibilities for achieving more effective milestones in building translation systems and machine translation tools [4], [7]. We are particularly interested in how well they can perform on machine translation tasks given the popular prompting methods across all these models adopted from their documentation.

LLMs, such as GPT-3 [4], [36], PaLM [7], BLOOM [51], and Llama [52] are trained to predict the next word by using the previous context. During unsupervised pre-training, a language model develops a broad set of pattern recognition abilities. It then uses these abilities at inference time to rapidly recognize and adapt to the desired task [34]. The term "in-context learning" was first brought up by Brown [4] to describe a scenario where a pre-trained language model at inference time learns to replicate certain input-output text generation patterns without further fine-tuning. [34]

Instead of giving a direct prompt to the model for performing a given task, our input can be augmented with relevant examples to help the model adapt its output. The model is expected to learn the pattern in the data examples and make better predictions accordingly [11]. Therefore, the key concept of in-context learning is to learn from analogy [34]. Using techniques like in-context learning with demonstrations [4], [1], the translation performance could be further improved.

In our case with MT tasks, LLMs can achieve more efficient translations by adapting their output to the terminology and style used in previously approved translation pairs, incorporating the philosophy of example-based and statistical MT [35], [31]; However, sometimes by using a purpose-based prompting framework the efficiency outperforms the n-shot method which is identical to the findings of Hendy [21].

Previous researchers have investigated MT with neural language models through few-shot in-context learning [54], as well as in zero-shot setting [59]. Other researchers have implied LLMs for generating synthetic domain-specific data for MT domain adaptation [33]. Recently, some of them [1] [63] have confirmed the importance of in-context example selection for the quality of MT with LLMs.

In this report, we provide a preliminary study of LLMs on MT, and we focus on some aspects:
• Translation Prompting Methods: LLMs use prompts as guidance to employ their translation capabilities accordingly. The style, arrangement, and generally the configuration of prompts affect the quality of outputs. For instance, the way our source or target language information is described matters in multilingual machine translation models. This is usually solved by attaching language tokens [27], [13].

• Multilingual Translation: LLMs are efficiently covering several languages, as a format of a unified multilingual MT model. Thus, we are curious about how they perform on different language pairs, considering they are different in terms of their family branch, the amount of their data engaged in training the model, and the direction in which we sort our source and target language. We use all combinations and directions of English, Persian, and Russian.

• Evaluation and Conclusion Based on both Automatic Evaluation metrics & Human Evaluation: We evaluated the translations based on multiple linguistics factors and metrics, analyzed and categorized the errors, and scored the translation outputs based on the severity of these errors. In addition, we explore the potential of our LLMs respectively, compared with modern neural metrics like COMET [44], BLEU [38], and Chrf++ [40].

By evaluating the translation among three selected languages on our test sets, we find that PaLM2 performs competitively on both high-resource languages like English or Russian as well as more under-resourced languages like Persian. GPT4 despite amazing results on high-resource languages, lags behind on a more distant language.

For language pairs that are both low-resource and from different families the performance gap can be further enlarged [58] Considering that parallel data between two distant languages is often scarce [13], [58] to be used in the n-shot setting.

Our findings align with Previous researchers [23], [21] [39] on translation tasks; they have found that an LLM like ChatGPT performs competitively with commercial translation products (e.g., Google Translate and Microsoft Translator) on high-resource languages, but proved to have fewer capabilities for low-resource and distant languages. Furthermore, PaLM could best perform translation prompt frameworks and analyze the examples given through in-context learning in an n-shot setting.

Furthermore, Results from [48] on PaLM-540B show a higher performance; We hypothesize that this is due to the large multilingual data proportion in its training data, which is 78% English and 22% for other languages [48], while the GPT data proportion is only 7% non-English [4].

We hypothesize that this may be due to the fact that this model can further improve its bilingual abilities as its training has involved multilingual data and can further improve this ability as it regulates and applies it more often; we also hypothesize that this capacity has improved analytical linguistics abilities as this model best handled the data sets which were given as examples and applied prompts regarding style and intonation more effectively compare to models with less multilingual abilities.

• The most accuracy and fluency were observed in the prompt enhanced zero-shot scenario rather than in n-shot scenario, in which we need to adjust our explanation settings according to the task given, complexity of the text, etc. our results show it has less stability and consistency when in-context learning shots are attached to them. Combining the two also didn't show significant improvements compare to its zero-shot setting.

• Despite their good output, we argue that LLMs are NOT a stable MT tool and there's ahigh possibility to generate multiple linguistics, literary errors, and hallucinations; this statement is highly dependent on the model we choose, the language direction, and the quality and relevancy of our prompts. It is NOT advisable to combine multiple translations into a single query input, as chatbots have a preference for former translations.

• The remainder of this report is designed as follows. We propose the evaluation settings and comparative results in Section 2, then through Section 3, we conclude our results, and in section 4 we highlight several potential issues that researchers' linguists and translators should be aware of when using LLMs as an MT tool.

## 2  Experimental Setup

## 2.1 Data Sets

For the purpose of multilingual translation and in-context learning, we evaluate the performance of the models on multilingual data; we evaluate our LLM-derived translation systems on the prominent and verified published versions of identical popular literary works. We tested all our samples through official APIs.

For the Persian language we used "My Uncle Napoleon" by Iraj Pezeshkzad, for the Russian language "The Beggar Boy at Christ's Christmas Tree" by Fyodor Dostoyevsky, and for the English language we used "The Tell-Tale Heart" by Edgar Allan Poe.

Regarding the importance of references for evaluating our translation outputs from LLMs, we use the best respective published translations of each of these works, that have been long enough evaluated by professionals within years. Figure 1 lists the information on these test sets. However, obtaining the translation results is a tough procedure, and is deeply time-consuming since it must be interacted with manually and cannot respond to large batches. Thus, we randomly sample 15-50 sentences from each set for evaluation.

For the n-shot technique, we implement examples, preferably from the same work not to interrupt with the intonation, style, etc.

| Test Set | Direction | Domain | Test set size Words |
|---|---|---|---|
| Persian culture Literary work | (Fa⇒En) (fa⇒Ru) | Literary- General | 255 |
| English culture Literary work | (En⇒ Fa) (En⇒ Ru) | Literary- General | 265 |
| Russian culture Literary work | (Ru⇒Fa) (Ru⇒En) | Literary- General | 161 |

Fig. 1. Information of Adopted Test Set

## 2.2 LLM Models

We compare a number of LLMs namely GPT 3.5, 4, PaLM2, meta-llama/Llama-2-70b, Claude 2, and Perplexity ai + Copilot. So far, we experiment with language pair combinations of English, Russian, and Persian. By default, the results in this report come from the versions on 2023.10.16. For new results, we will mark the updated version information correspondingly.

## 2.3 Evaluation System:

*2.3.1 Automatic Evaluation Metrics.* Previous researcher [15] recommends using neural network-based metrics, as they have demonstrated to have high correlation with human evaluation results. Hence, we adopt the mostly used BLEU score [38] as our primary metric and also report chrF++ [40], and COMET [44].

*2.3.2 Human Evaluation Method.* Each year thousands of human judgments are released to evaluate the quality of MT systems to determine potential algorithms and techniques as the new state-of-the-art. In a typical scenario, human judges evaluate a system's output by comparing it to a source as a reference translation. Then, they score their primary hypothesis according to a set of criteria, including fluency and adequacy [61]; or rank a set of hypotheses according to an order of preferences [55],[5][18]. This task is highly challenging, first and foremost as there are no standard guidelines defined for it. Secondly, finding a framework consisting of the most common errors is quite time-consuming and requires consistent efforts. Thirdly, it is dependent on the evaluator's background, level of literacy in languages, and bilingual/ multilingual abilities to distinguish linguistics and translation aspects in a text. [5]. As a result, evaluations suffer from low inter- and intra-annotator agreements [53], [49]. Yet, as Sanders [45] argue, using human judgments is essential to the progress of MT systems because: (i) automatic translations as our MT outputs are produced for a human audience; and (ii) human understanding of the real world allows to assess the importance of the errors made by MT systems. [19]

Vilar [56] proposed that sometimes, the interpretation of automatic metric scores turns out to be unclear; therefore, error classification and analysis by humans is essential. [12] Turian [53] also insisted on the importance of human judgment. Several experts disagreed with the statement that automatic metrics are demonstrated to have a good correlation with human judgments [10], [6][12]. In this regard, Sennrich [46] also pointed out that BLEU only focuses on precision and does not consider syntactic structures and grammar.[12] Furthermore, another researcher [37], noted that BLEU scores are not efficient when it comes to the evaluation of NMT. [12]

Llitjós [14] who aimed to find an automation process for post-editing, was one of the first experts to present an error typology. [12] Their proposed classification served as a model for Vilar work [56], regarding human evaluation of SMT. Both of these classifications are indeed very similar, with three categories in common ("missing word", "word order" and "incorrect words"). [12] Vilar [56] proposed a more comprehensive typology, which contained more sub-categories, and as a result a more precise identification of errors. For instance, the sub-category namely "sense" which belongs to "word order" has been divided into two categories namely "wrong lexical choice" and "incorrect disambiguation". Daems [9] in their work, proposed a typology with similar categories, even though they were originally aiming to quantify the post-editing process. Although the general classification appears to be different, these frameworks share several common features (for instance the "lexicon" category in the research by Daems [9] is similar to that of "wrong lexical choice" in the other work [56][12].

In our work, in addition to neural network-based metrics we implemented human evaluation, and prioritize that indeed; from the beginning of our research, we adopt test sets in a setting that allows a consistent human evaluation for our results to formulate as precise as possible. All our MT outputs from LLMs were investigated by a couple of translators and linguists efficient in our 3 selected language combinations; despite other similar categorizations in the past research, by adjusting our test sets we decided we needed to build a typology specified for the purpose of our research; and in fact, it shares several common aspects and features with previously approved typologies. most common errors were identified and categorized into 8 groups for each language direction as presented in the list below:

## Error Categories

**Fa to Ru:**
1. Literal Translation
2. Literal Meaning
3. Word Choice
4. Word Omission
5. Syntax
6. Grammar
7. Content Deviation

**Ru to Fa:**
1. Incorrect Order
2. Addition
3. Word Choice
4. Syntax
5. Word Omission
6. Incorrect Punctuation
7. Content Deviation

**Eng to Fa:**
1. Incorrect Punctuation
2. Incorrect Word Collocation
3. Grammar
4. Literal Translation
5. Literal Meaning
6. Misspelling
7. Addition

**Eng to Ru:**
1. Word Omission
2. Content Deviation
3. Addition
4. literal meaning
5. Grammar
6. Incorrect Punctuation
7. Incorrect Translation of time

**Fa to Eng:**
1. Repetition
2. Wordiness
3. Word choice
4. Word omission
5. Syntax & order
6. Idiomatic

**Ru to Eng:**
1. Literal translation
2. Repetition & Addition
3. Word choice
4. Word omission
5. Syntax
6. Grammar
7. Content deviation

Fig. 2. Information of Adopted Test Set

## 3 Experimental Setup

Compared to traditional machine translation systems, LLMs like ChatGPT can incorporate additional information, like human interactions, through input prompts [10]. The performance of LLMs has shown significant enhancements by using in-context learning on our test input by providing a few labeled examples (prompts) [4]. This few-shot paradigm has demonstrated strong performance across multiple natural language processing (NLP) tasks [36], [18][60], [7]. Other researchers have also released a series of recent works regarding in-context learning on machine translation (MT) with rather mixed results and various ways of shot selection. Zhang in their work [63] used GLM-130B and proposed consistent but rather low correlation when using carefully selected shots compared to using random shots in the performance of MT. They use different features to show varying levels of this correlation, and its effect on the performance. In the same spirit, Vilar [54] used different prompt selection schemes with the PaLM- 540B model, and concluded that using input-relevant examples is not necessarily better than using random shots; they rather pointed out the importance of high-quality examples for the shots. In the same vein, Agrawal [1] used a much smaller model XGLM-7.5B, and employed multiple selection criteria for their selected shots. They showed that a combination of retrieval and task metrics is consistently better than the random baseline across different translation directions.

In this work, we are curious to compare the performance of the raw n-shot method using a translation-tailored prompting framework, and finally combine these two settings into one, and execute our prompting frameworks while incorporating them into our nominated shots.

### 3.1    Prompting Strategies:

For designing a set of prompts, in order to trigger the machine translation ability of the selected LLMs, we used two of the most prominent methods and combined them as well as our final prompting setup. As illustrated in Figure 2, our first method is the n-shot prompting, and for the second one, we used a well-known standard prompting framework specialized for translation and localization purposes adopted from "Keyword Everywhere" which provides a set of pre-tested frameworks and templates for various purposes with lots of flexibilities. We used the same template and prompt for each testing stage across all our language directions and language models.

Last but not least, we combined the n-shot method and prompting frameworks to see whether it affects the level of efficiency using our selected evaluation methods. Thus, we summarized our candidate prompts as presented in Figure 2, where [SRC] and [TGT] represent our source and target languages in our experiment.

| Simple Zero to Few-Shot Prompts | |
|---|---|
| Tp1 for zero-shot | Translate the following prose from [SRC] to [TGT]: "txt" |
| Tp2 for one-shot | Example: [SRC] prose: "txt"<br>[TGT] translation:"txt"<br>translate the following prose from [SRC] to [TGT]: "txt" |
| Tp3 for two-shot | Example NO.1: [SRC] prose:"txt"<br>[TGT] translation:"txt"<br>Example NO.2: [SRC] prose:"txt"<br>[TGT] translation:"txt"<br>Translate the following prose from [SRC] to [TGT]: "txt" |
| Tp3  for few-shot | Example NO.1: [SRC] prose:"txt"<br>[TGT] translation:"txt"<br>Example NO.2: [SRC] prose:"txt"<br>[TGT] translation:"txt"<br>Example NO.3: [SRC] prose:"txt"<br>[TGT] translation:"txt"<br>translate the following prose from [SRC] to [TGT]: "txt" |

Fig. 3. Candidate translation prompts

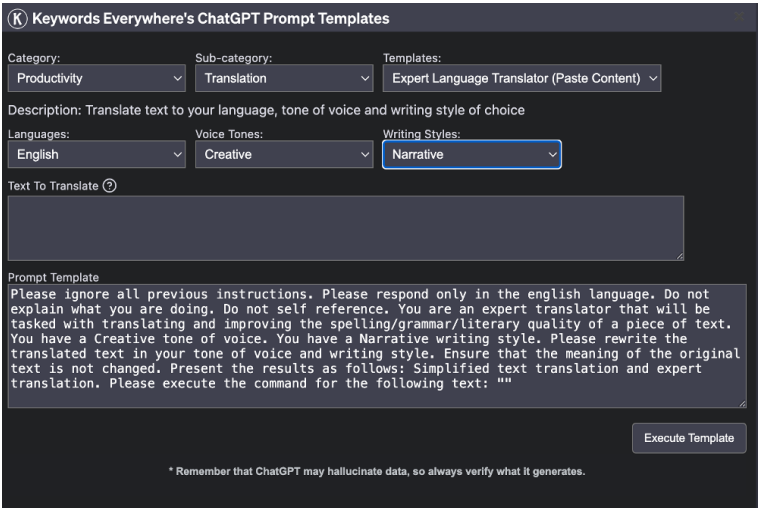

Fig. 4. Candidate translation prompting framework

We compared the three different candidate prompt categories on the Persian to English (Fa⇒En), Persian to Russian (Fa⇒Ru), English to Persian (En⇒ Fa), English to Russian (En⇒ Ru), Russian to Persian (Ru⇒ Fa), and Russian to English (Ru⇒En) translation tasks with the test set from our data as described. Our selected LLMs include: GPT 3.5, GPT4, PaLM2, Claude-2-100k, meta-llama/Llama-2-70b-, Perplexity AI (incorporating Copilot).

## 3.2 N-Shot Translation

*3.2.1 Explanation.* In this section, we explore the performance of each of our selected LLMs for each of the language combinations separately, and according to the number of the examples/ shots we used to provide context in our prompts.

N-shot prompting performance varies greatly across the number of our shots and also varies greatly depending on the direction of our data set. We start with zero-shot prompting and explore the effect of increasing the given shots. Depending on how to describe MT and partially inspired by prior studies [4], [7], [60], we compared the N-shot method on our 6 different directions in the tables below, the results for each of the directions have been sorted, as all the results across all the models vary depending on the selected languages, the direction in which they are sorted.

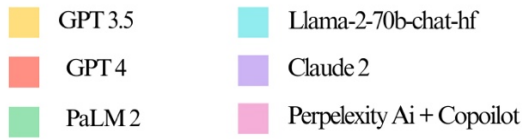

Fig. 5. Color Guides

(FA⇒RU)

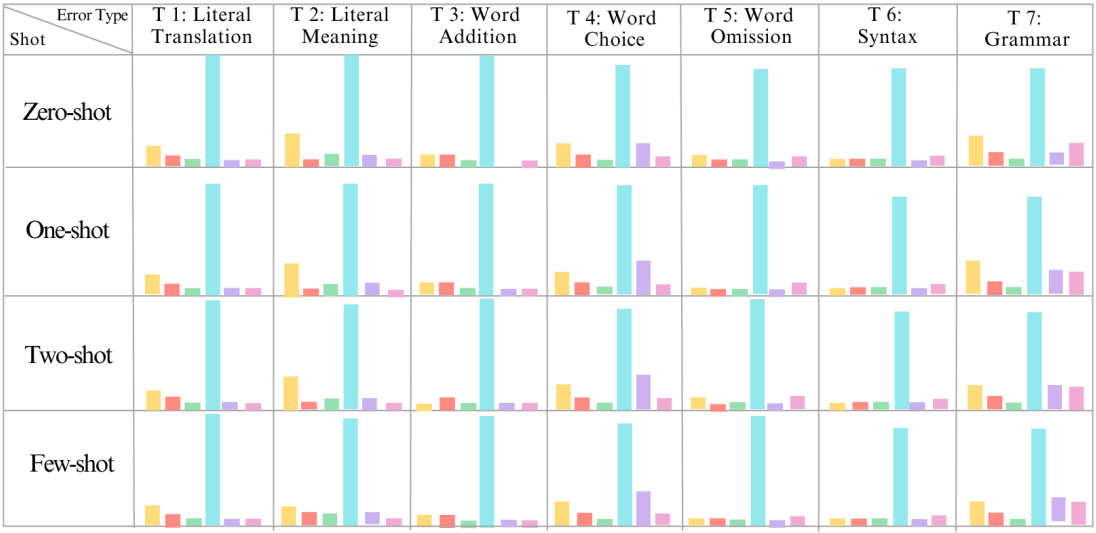

Fig. 6. Human error analysis on n-shot method, Persian to Russian

(RU⇒FA)

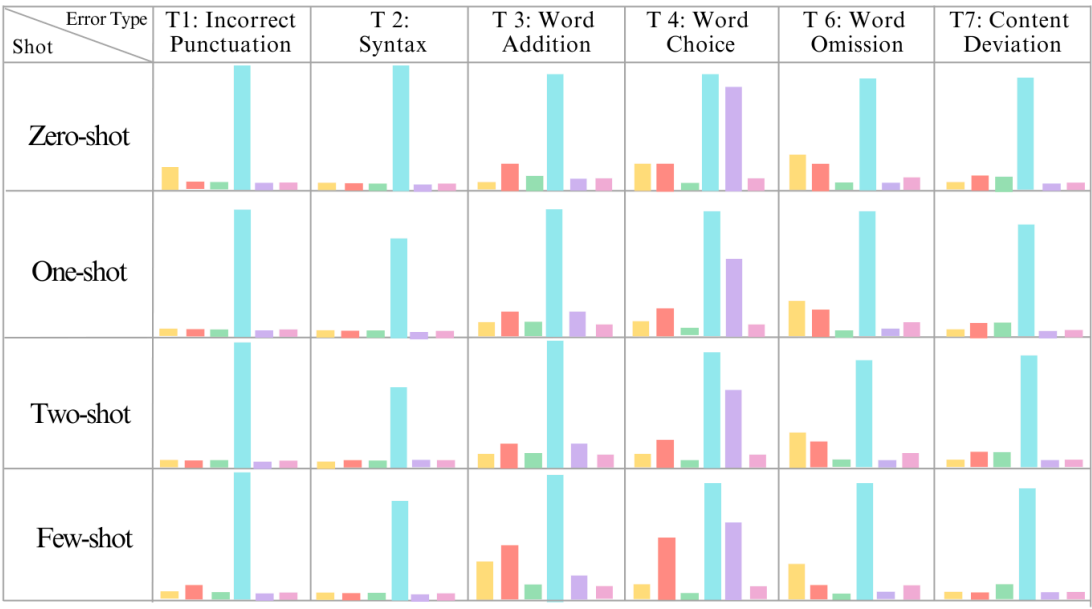

Fig. 7. Human error analysis on n-shot method, Russian to Persian

(EN⇒FA)

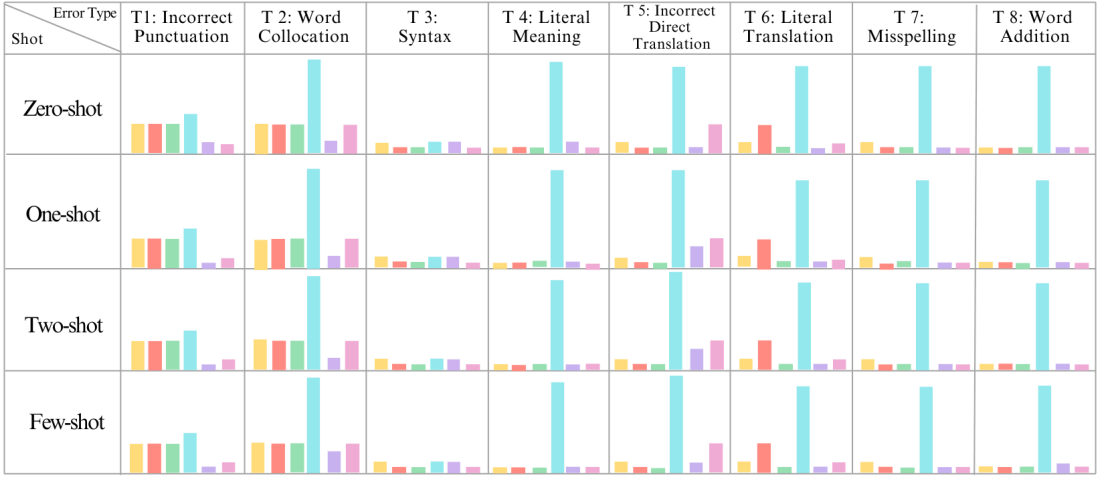

Fig. 8. Human error analysis on n-shot method, English to Persian

(FA⇒EN)

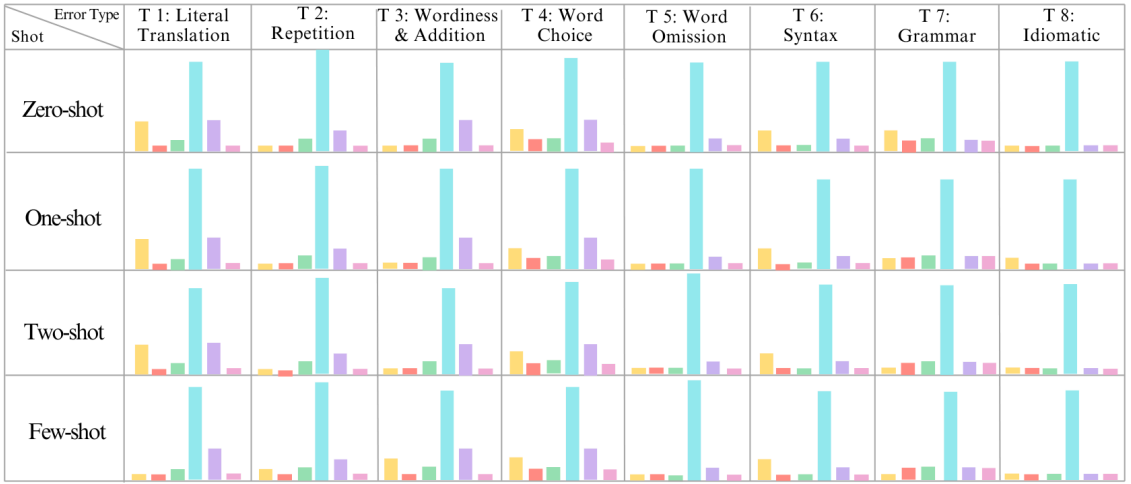

Fig. 9. Human error analysis on n-shot method, Persian to English

(RU⇒EN)

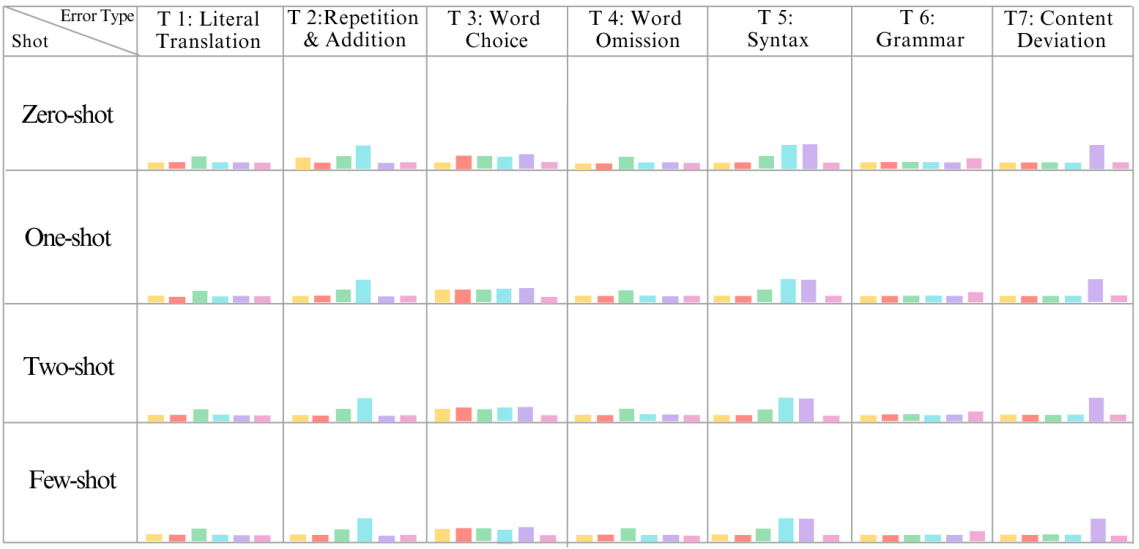

Fig. 10. Human error analysis on n-shot method, Russian to English

(EN⇒RU)

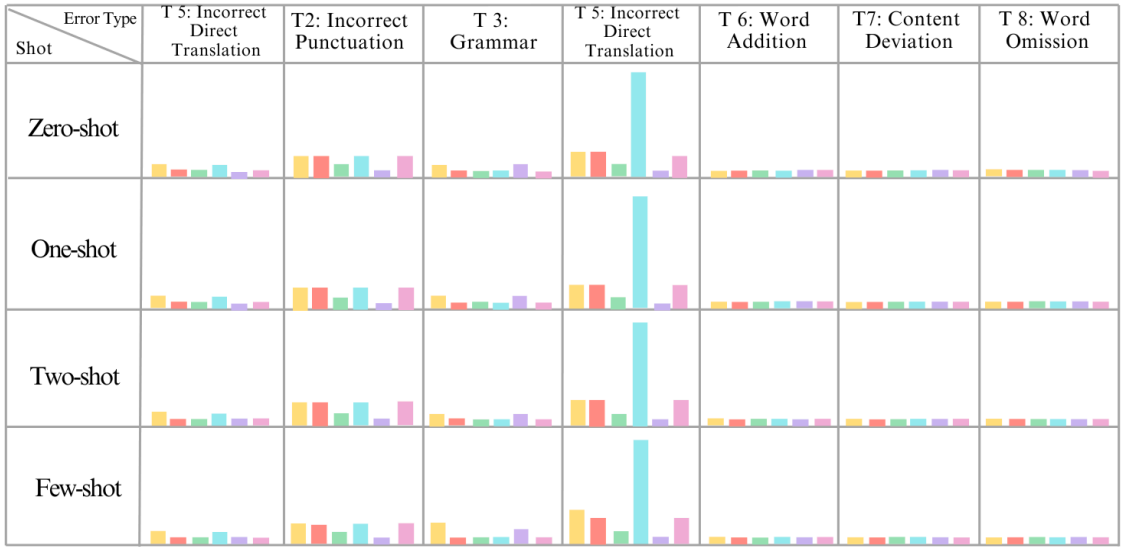

Fig. 11. Human error analysis on n-shot method, English to Russian

### 3.2.2 Results based on Linguistics Errors.
In the first set of the tables, for each direction, you can distinguish the amounts of errors for each of the shots, and compare the performance of every model together

| Fa to Ru | | 0 to 1 shot | 1 to 2 Shot | 2 to Few Shot | |
|---|---|---|---|---|---|
| | GPT 3.5 | | | | T1 |
| | | | | Decrease↓ (by 1) | T2 |
| | | | Decrease↓ (by 1) | Increase↑ (by 1) | T3 |
| | | | | | T4 |
| | | Decrease↓ (by 1) | Increase↑ (by 1) | Decrease↓ (by 1) | T5 |
| | | | | | T6 |
| | | | Decrease↓ (by 1) | | T7 |
| | GPT 4 | | | | T1 |
| | | | Increase↑ (by 1) | | T2 |
| | | | | | T3 |
| | | | | | T4 |
| | | | | | T5 |
| | | | | | T6 |
| | | | | | T7 |
| | PaLM | Remained Fixed across all shots and errors | | | |
| | Llama | Translation rejected not qualified for evaluation | | | |
| | Claude | | | | T1 |
| | | | | | T2 |
| | | | | | T3 |
| | | Decrease↓ (by 1) | | | T4 |
| | | | | | T5 |
| | | | | | T6 |
| | | Increase↑ (by 1) | | | T7 |
| | Perplexity | Remained Fixed across all shots and errors | | | |
| Results: | | | | | |
| GPT 3.5 shows the most instability. | | | | | |
| Most sufficient model is PaLM, after that GPT4. | | | | | |
| T4 and T7 were two most dominant errors in this direction. | | | | | |
| Overall, in GPT 3.5 despite inefficiency, we observe, increasing shots has lessen the errors. | | | | | |

| Ru to fa | | 0 to 1 shot | 1 to 2 Shot | 2 to Few Shot | |
|---|---|---|---|---|---|
| | GPT 3.5 | Decrease↓ (by 2) | | | T1 |
| | | | | | T2 |
| | | Increase↑ (by 1) | | Increase↑ (by 2) | T3 |
| | | Decrease↓ (by 1) | | | T4 |
| | | | | | T5 |
| | | | | | T6 |
| | GPT 4 | | | Increase↑ (by 1) | T1 |
| | | | | | T2 |
| | | | | Increase↑ (by 2) | T3 |
| | | | | Increase↑ (by 3) | T4 |
| | | | | Decrease↓ (by 1) | T5 |
| | | | | Decrease↓ (by 1) | T6 |
| | PaLM | Remained Fixed across all shots and errors | | | |
| | Llama | Translation rejected not qualified for evaluation | | | |
| | Claude | | | | T1 |
| | | | | | T2 |
| | | Increase↑ (by 1) | | | T3 |
| | | Decrease↓ (by 2) | | | T4 |
| | | | | | T5 |
| | | | | | T6 |
| | Perplexity | Remained Fixed across all shots and errors | | | |
| Results: | | | | | |
| GPT 4 showed the most instability, after that GPT 3.5. | | | | | |
| For this direction PaLM and then Perplexity were the best options. | | | | | |
| T4,5 and 3 were the most common error types. | | | | | |
| As you can see increasing shots not necessarily output less errors. We observe instability in that as well. | | | | | |

Fig. 12. The effect of adding shots, Fa to Ru.   Fig. 13. The effect of adding shots, Ru to Fa

| Eng to fa | | 0 to 1 shot | 1 to 2 Shot | 2 to Few Shot | |
|---|---|---|---|---|---|
| | GPT 3.5 | Remained Fixed across all shots and errors | | | |
| | GPT 4 | Remained Fixed across all shots and errors | | | |
| | PaLM | Remained Fixed across all shots and errors | | | |
| | Llama | Translation rejected not qualified for evaluation | | | |
| | Claude | Decrease↓ (by 1) | | | T1 |
| | | | | Increase↑ (by 1) | T2 |
| | | | | | T3 |
| | | Decrease↓ (by 1) | | | T4 |
| | | Increase↑ (by 2) | | Decrease↓ (by 1) | T5 |
| | | | | | T6 |
| | | | | | T7 |
| | | | | Increase↑ (by 1) | T8 |
| | Perplexity | Remained Fixed across all shots and errors | | | |
| Results: | | | | | |
| Claud shows the most Instability. | | | | | |
| Best models are PaLM, and after that GPT4. | | | | | |
| T1, 2 and 5, are the most common error types in this direction. | | | | | |
| Shots did not have any effect on increasing or decreasing the errors in any of the models, and with the Claude the positive and negative effects are equal. | | | | | |

| Fa to Eng | | 0 to 1 shot | 1 to 2 Shot | 2 to Few Shot | |
|---|---|---|---|---|---|
| | GPT 3.5 | | | Decrease↓ (by 3) | T1 |
| | | | | Increase↑ (by 1) | T2 |
| | | | | Increase↑ (by 2) | T3 |
| | | | | | T4 |
| | | | | | T5 |
| | | | | | T6 |
| | | Decrease↓ (by 1) | Decrease↓ (by 1) | | T7 |
| | | Increase↑ (by 1) | Decrease↓ (by 1) | | T8 |
| | GPT 4 | Remained Fixed across all shots and errors | | | |
| | PaLM | Remained Fixed across all shots and errors | | | |
| | Llama | Translation rejected not qualified for evaluation | | | |
| | Claude | Remained Fixed across all shots and errors | | | |
| | Perplexity | Remained Fixed across all shots and errors | | | |
| Results: | | | | | |
| Perplexity and GPT4 were slightly better than PaLM. | | | | | |
| There is no distinct common error type in this direction; errors are distributed across all of them more or less equally; probably T4 | | | | | |
| Overall n-shot has showed efficiency with GPT 3.5, which is the only model with changes as a result of adding shots. | | | | | |

Fig. 14. The effect of adding shots, Eng to Fa.   Fig. 15. The effect of adding shots, Fa to Eng

| Ru to Eng | | 0 to 1 shot | 1 to 2 Shot | 2 to Few Shot | |
|---|---|---|---|---|---|
| | GPT 3.5 | | | | T1 |
| | | Decrease↓ (by 1) | | | T2 |
| | | Increase↑ (by 1) | | | T3 |
| | | | | | T4 |
| | | | | | T5 |
| | | | | | T6 |
| | | | | | T7 |
| | GPT 4 | Remained Fixed across all shots and errors | | | |
| | PaLM | Remained Fixed across all shots and errors | | | |
| | Llama | Remained Fixed across all shots and errors | | | |
| | Claude | Remained Fixed across all shots and errors | | | |
| | Perplexity | Remained Fixed across all shots and errors | | | |
| Results: | | | | | |
| Overall, in this direction, the least number of errors occurred. | | | | | |
| Claude and Llama fall behind other models. | | | | | |
| Perplexity and GPT4 were again slightly better than PaLM. | | | | | |
| Despite of the fact that the errors are almost equally distributed, T5 and T3 are most likely to happen. | | | | | |

| Eng to Ru | | 0 to 1 shot | 1 to 2 Shot | 2 to Few Shot | |
|---|---|---|---|---|---|
| | GPT 3.5 | | | | T1 |
| | | | | | T2 |
| | | | | Increase↑ (by 1) | T3 |
| | | | | | T4 |
| | | | | | T5 |
| | | | | | T6 |
| | | | | | T7 |
| | GPT 4 | Remained Fixed across all shots and errors | | | |
| | PaLM | Remained Fixed across all shots and errors | | | |
| | Llama | Remained Fixed across all shots and errors | | | |
| | Claude | Remained Fixed across all shots and errors | | | |
| | Perplexity | Remained Fixed across all shots and errors | | | |
| Results: | | | | | |
| The most efficient model is PaLM, after that Claude. | | | | | |
| T4 and T2 are the most common error types. | | | | | |
| Increasing Shots almost did not affect the number of errors at all. | | | | | |

Fig. 16. The effect of adding shots, Ru to Eng    Fig. 17. The effect of adding shots, Eng to Ru

## 3.3 Standard Prompt Enhanced Framework on Zero-shot & N-shot

*3.3.1 Explanation.* In this section, we explore the performance of each our LLMs for each of language combinations separately, according to the adopted tailored prompting framework for this setting. In zero-shot case, we proposed the impact of the frameworks on the performance solely, then in the one, two, and few-shot setting we also incorporated our shots combined with the prompting framework. In the tables below, the results for each of the directions has been sorted, as all the results across all the models vary depending on the selected languages, and the direction in which they are sorted. In the first set of the tables, for each direction you can distinguish the error rates for each of the shots, and compare the performance of models together. Then in the second set of the tables the transition rate of these errors has been tracked for better understanding the effect of adding shots on the number of the errors and their types.

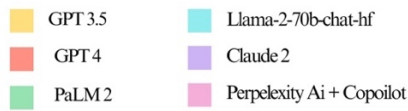

Fig. 18. Color guides

(FA⇒RU)

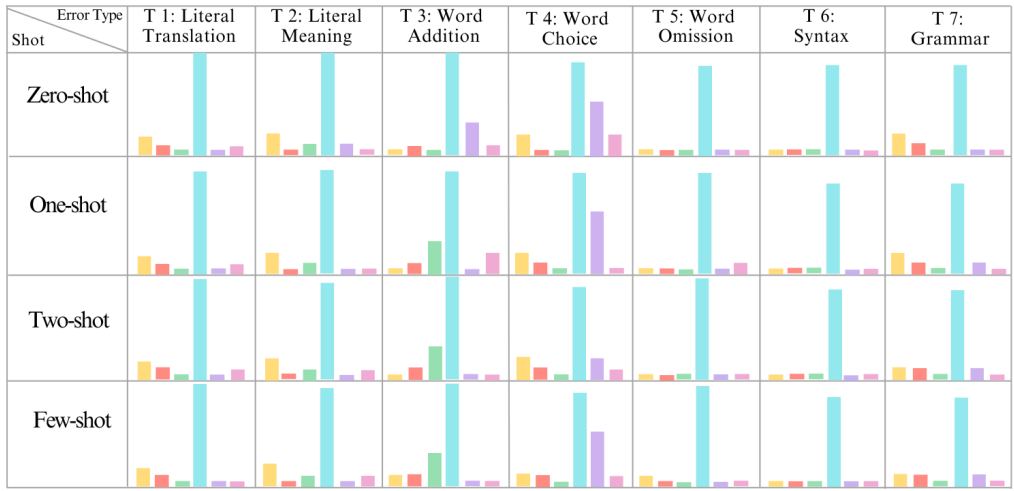

Fig. 19. Human error analysis on Prompt enhanced n-shot method, Persian to Russian

(RU⇒FA)

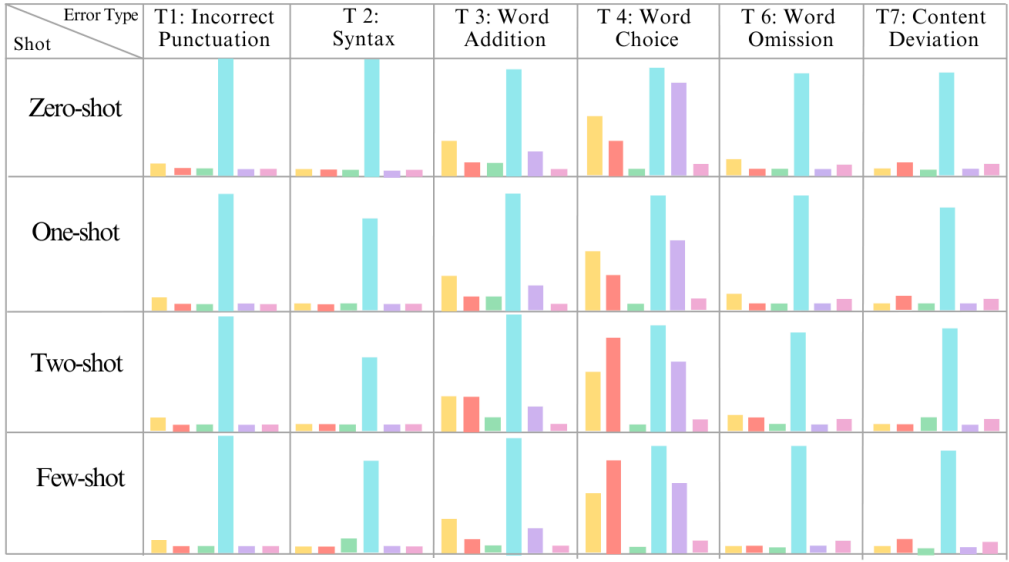

Fig. 20. Human error analysis on Prompt enhanced n-shot method, Russian to Persian

(EN⇒FA)

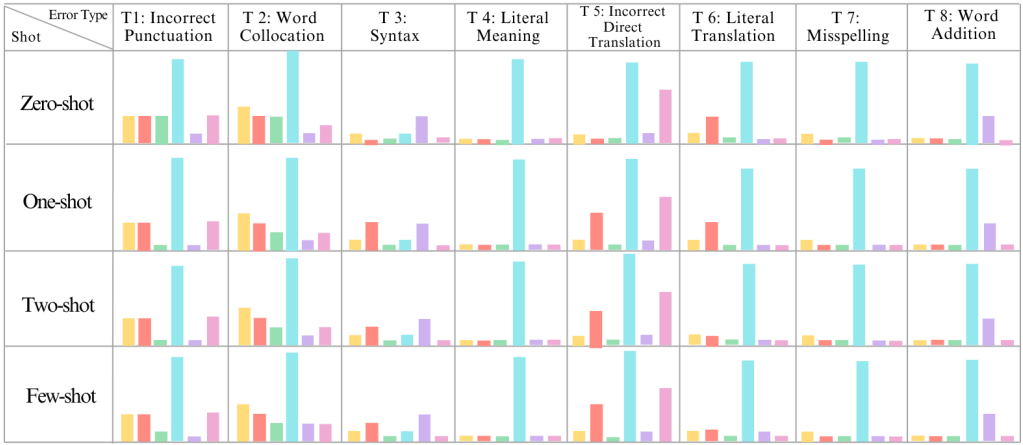

Fig. 21. Human error analysis on Prompt enhanced n-shot method, English to Persian

(FA⇒EN)

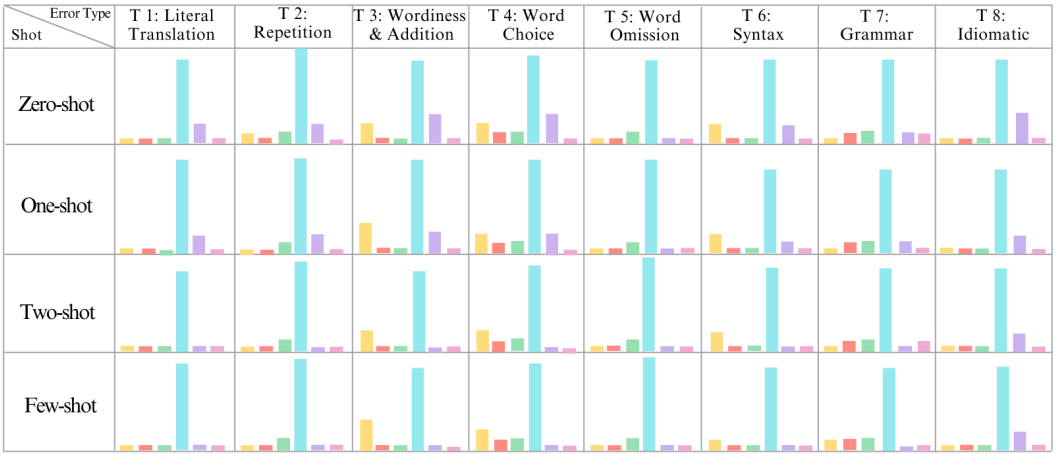

Fig. 22. Human error analysis on Prompt enhanced n-shot method, Persian to English

(RU⇒EN)

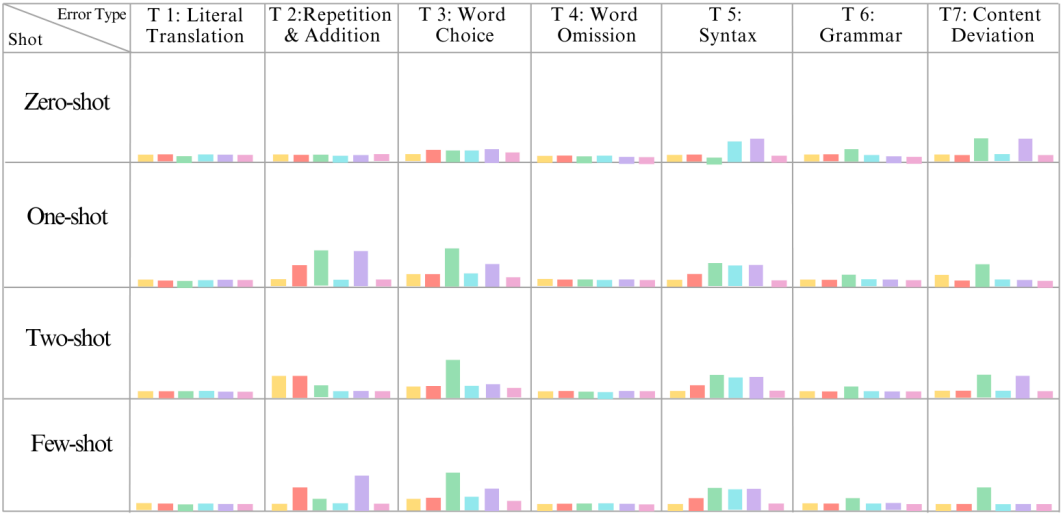

Fig. 23. Human error analysis on Prompt enhanced n-shot method, Russian to English

(EN⇒RU)

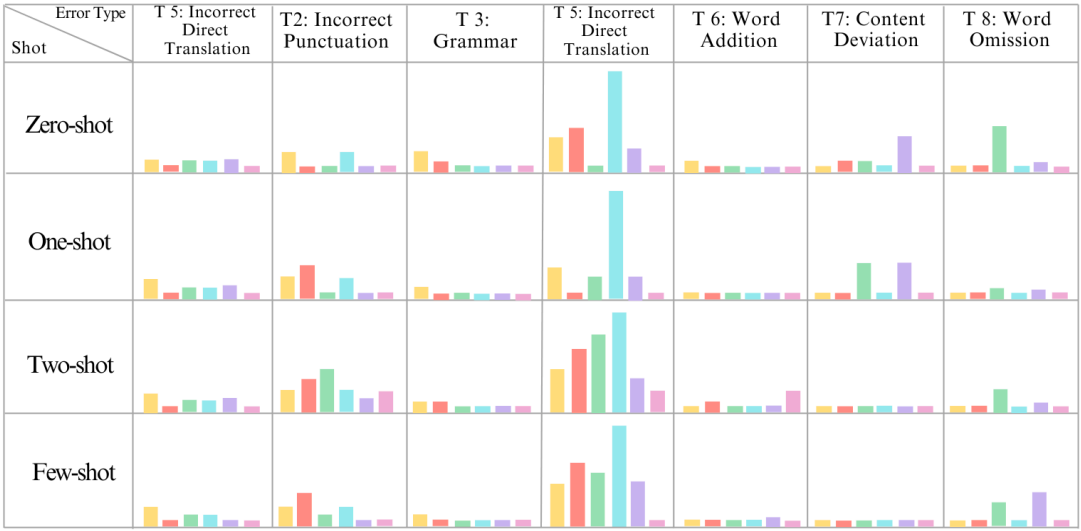

Fig. 24. Human error analysis on Prompt enhanced n-shot method, English to Russian

*3.3.2 Results based on Linguistics Errors.* In the first set of the tables, for each direction, you can distinguish the error rates for each of the shots, and compare the performance of every model together

| Fa to Ru | | 0 to 1 shot | 1 to 2 Shot | 2 to Few Shot | | Ru to Fa | | 0 to 1 shot | 1 to 2 Shot | 2 to Few Shot | |
|---|---|---|---|---|---|---|---|---|---|---|---|
| | GPT 3.5 | | | | T1 | | GPT 3.5 | | | | T1 |
| | | | | | T2 | | | | | | T2 |
| | | | | Increase↑ (by 1) | T3 | | | | | | T3 |
| | | | | Decrease↓ (by 1) | T4 | | | | | | T4 |
| | | | | Increase↑ (by 1) | T5 | | | | | Decrease↓ (by 1) | T5 |
| | | | | | T6 | | | | | | T6 |
| | | | Decrease↓ (by 1) | | T7 | | | | | | |
| | GPT 4 | | | | T1 | | GPT 4 | | | | T1 |
| | | | | | T2 | | | | | | T2 |
| | | | | | T3 | | | | Increase↑ (by 1) | Decrease↓ (by 1) | T3 |
| | | Increase↑ (by 1) | | | T4 | | | | Increase↑ (by 8) | | T4 |
| | | | | | T5 | | | | Increase↑ (by 1) | Decrease↓ (by 1) | T5 |
| | | | | | T6 | | | | Decrease↓ (by 1) | Increase↑ (by 1) | T6 |
| | | | | | T7 | | | | | | |
| | PaLM | | | | T1 | | PaLM | | | | T1 |
| | | | | | T2 | | | | | Increase↑ (by 1) | T2 |
| | | Increase↑ (by 3) | | | T3 | | | | | Decrease↓ (by 1) | T3 |
| | | | | | T4 | | | | | | T4 |
| | | | | | T5 | | | | | | T5 |
| | | | | | T6 | | | | Increase↑ (by 1) | Decrease↓ (by 1) | T6 |
| | | | | | T7 | | | | | | |
| | Llama | Translation rejected not qualified for evaluation | | | | | Llama | Translation rejected not qualified for evaluation | | | |
| | Claude | | | | T1 | | Claude | | | | T1 |
| | | Decrease↓ (by 1) | | | T2 | | | | | | T2 |
| | | Decrease↓ (by 3) | | | T3 | | | | | | T3 |
| | | | Decrease↓ (by 4) | Increase↑ (by 4) | T4 | | | Decrease↓ (by 2) | | | T4 |
| | | | | | T5 | | | | | | T5 |
| | | | | | T6 | | | | | | T6 |
| | | Increase↑ (by 1) | | | T7 | | | | | | |
| | Perplexity | | | Decrease↓ (by 1) | T1 | | Perplexity | Remained Fixed across all shots and errors | | | |
| | | | Increase↑ (by 1) | | T2 | | | | | | |
| | | Increase↑ (by 1) | | Decrease↓ (by 2) | T3 | | | | | | |
| | | Decrease↓ (by 2) | Increase↑ (by 1) | | T4 | | | | | | |
| | | Increase↑ (by 1) | Decrease↓ (by 1) | | T5 | | | | | | |
| | | | | | T6 | | | | | | |
| | | | | | T7 | | | | | | |

**Results:**
Perplexity showed the most instability.
Despite the Instability of Perplexity, this model and GPT4 were the bests in this direction and the prompting method taken here.
Claude shows significant weakness with the T4.
The method taken not necessarily improves efficiency.

**Results:**
GPT4 and PaLM showed the most instability.
PaLM and Perplexity are the best models in this direction and with this method.
T4 is the error category, which most likely to happen in this method and direction.
The worst outputs were from GPT4.
PaLM has changed style to literary more than needed; also, in few-shot it had digested the text and has written an inference instead of giving a proper translation but the content has not changed at all.

Fig. 25. The effect of adding shots, Fa to Ru    Fig. 26. The effect of adding shots, Ru to Fa

| Eng to Fa | | 0 to 1 shot | 1 to 2 Shot | 2 to Few Shot | |
|---|---|---|---|---|---|
| | GPT 3.5 | Remained Fixed across all shots and errors | | | |
| | GPT 4 | | | | T1 |
| | | | | | T2 |
| | | Increase↑ (by 3) | Decrease↓ (by 1) | | T3 |
| | | Increase↑ (by 4) | | | T4 |
| | | Decrease↓ (by 2) | | | T5 |
| | | | | | T6 |
| | | | | | T7 |
| | | | | | T8 |
| | PaLM | Decrease↓ (by 3) | | Increase↑ (by 1) | T1 |
| | | Decrease↓ (by 1) | | | T2 |
| | | | | | T3 |
| | | | | | T4 |
| | | | | | T5 |
| | | | | | T6 |
| | Llama | Translation rejected not qualified for evaluation | | | |
| | Claude | Decrease↓ (by 1) | | | T1 |
| | | | | Increase↑ (by 1) | T2 |
| | | | | | T3 |
| | | | | | T4 |
| | | | | | T5 |
| | | | | | T6 |
| | | | | | T7 |
| | Perplexity | Remained Fixed across all shots and errors | | | |

Results:
Most instability occurred with GPT4.
Most stability with Perplexity and GPT3.5
PaLM is the best model here.
T2,1 and 5 are the most common error types here.
PaLM again made the style to literary in prompt enhanced.

| Fa to Eng | | 0 to 1 shot | 1 to 2 Shot | 2 to Few Shot | |
|---|---|---|---|---|---|
| | GPT 3.5 | | | | T1 |
| | | | | | T2 |
| | | Increase↑ (by 1) | Decrease↓ (by 1) | Increase↑ (by 1) | T3 |
| | | | | | T4 |
| | | | | | T5 |
| | | | | Decrease↓ (by 1) | T6 |
| | | | | Increase↑ (by 1) | T7 |
| | | | | | T8 |
| | GPT 4 | Remained Fixed across all shots and errors | | | |
| | PaLM | Remained Fixed across all shots and errors | | | |
| | Llama | Translation rejected not qualified for evaluation | | | |
| | Claude | | Decrease↓ (by 2) | | T1 |
| | | | Decrease↓ (by 2) | | T2 |
| | | Decrease↓ (by 1) | Decrease↓ (by 2) | | T3 |
| | | | | | T4 |
| | | | | | T5 |
| | | Decrease↓ (by 1) | Decrease↓ (by 1) | | T6 |
| | | | Decrease↓ (by 1) | | T7 |
| | | Decrease↓ (by 1) | | | T8 |
| | Perplexity | | | | T1 |
| | | | | | T2 |
| | | | | | T3 |
| | | | | | T4 |
| | | | | | T5 |
| | | | | | T6 |
| | | Decrease↓ (by 1) | Increase↑ (by 1) | Decrease↓ (by 1) | T7 |
| | | | | | T8 |

Results:
As shown in the chart, Claude shows the most instability, but the process has worked significantly efficient with this model and this setting and the method taken here.
Perplexity is the best model and after that GPT4.
Despite the fact that, errors were distributed almost equally, T4 seems to be the most common error type.

Fig. 27. The effect of adding shots, Eng to Fa    Fig. 28. The effect of adding shots, Fa to Eng

| Ru to Eng | | 0 to 1 shot | 1 to 2 Shot | 2 to Few Shot | |
|---|---|---|---|---|---|
| | GPT 3.5 | | | | T1 |
| | | | Increase↑ (by 2) | Decrease↓ (by 2) | T2 |
| | | Increase↑ (by 1) | | | T3 |
| | | | | | T4 |
| | | | | | T5 |
| | | | | | T6 |
| | GPT 4 | | | | T1 |
| | | Increase↑ (by 2) | | | T2 |
| | | | | | T3 |
| | | | | | T4 |
| | | Decrease↓ (by 1) | | | T5 |
| | | | | | T6 |
| | PaLM | | | | T1 |
| | | Increase↑ (by 3) | Decrease↓ (by 2) | | T2 |
| | | Increase↑ (by 3) | | | T3 |
| | | | | | T4 |
| | | Increase↑ (by 2) | | | T5 |
| | | | | | T6 |
| | | | | | T7 |
| | Llama | Remained Fixed across all shots and errors | | | |
| | Claude | | | | T1 |
| | | Increase↑ (by 3) | Decrease↓ (by 3) | Increase↑ (by 3) | T2 |
| | | Increase↑ (by 1) | Decrease↓ (by 1) | Increase↑ (by 1) | T3 |
| | | | | | T4 |
| | | | | | T5 |
| | | | | | T6 |
| | | Decrease↓ (by 2) | Increase↑ (by 2) | Decrease↓ (by 2) | T7 |
| | Perplexity | Remained Fixed across all shots and errors | | | |

Results:
Claude shows the most instability.
Perplexity is the best model and after that comes the GPT4.
T5, 2, and 7 are the most common error types.

| Eng to Ru | | 0 to 1 shot | 1 to 2 Shot | 2 to Few Shot | |
|---|---|---|---|---|---|
| | GPT 3.5 | Increase↑ (by 1) | | | T1 |
| | | | | | T2 |
| | | Decrease↓ (by 1) | | | T3 |
| | | | Increase↑ (by 1) | | T4 |
| | | Decrease↓ (by 1) | | | T5 |
| | | | | | T6 |
| | | | | | T7 |
| | GPT 4 | | | | T1 |
| | | Increase↑ (by 3) | | | T2 |
| | | Decrease↓ (by 1) | Increase↑ (by 1) | Decrease↓ (by 1) | T3 |
| | | Decrease↓ (by 4) | Increase↑ (by 6) | | T4 |
| | | | Increase↑ (by 1) | Decrease↓ (by 1) | T5 |
| | | Decrease↓ (by 1) | | | T6 |
| | | | | | T7 |
| | PaLM | | | | T1 |
| | | | Increase↑ (by 4) | Decrease↓ (by3) | T2 |
| | | | | | T3 |
| | | Increase↑ (by 2) | Increase↑ (by 5) | Decrease↓ (by 2) | T4 |
| | | | | | T5 |
| | | Increase↑ (by 2) | | | T6 |
| | | Decrease↓ (by 3) | Increase↑ (by 1) | | T7 |
| | Llama | Remained Fixed across all shots and errors | | | |
| | Claude | | | Decrease↓ (by 1) | T1 |
| | | | Increase↑ (by 1) | Decrease↓ (by 1) | T2 |
| | | | | | T3 |
| | | | Increase↑ (by 1) | Increase↑ (by 1) | T4 |
| | | | | Increase↑ (by 1) | T5 |
| | | Decrease↓ (by 3) | | | T6 |
| | | | | Increase↑ (by 2) | T7 |
| | Perplexity | Remained Fixed across all shots and errors | | | |

Results:
Best model is Perplexity.
T4 is the most common error type, and after that T2.

Fig. 29. The effect of adding shots, Ru to Eng    Fig. 30. The effect of adding shots, Eng to Ru

### 3.3 Overall Results

1.     The efficiency of the model not just was relevant and dependent to the language direction but also was depends on the target language; if the target language in our desired language direction was a high-resource language like English the efficiency was much more higher comparing to the other way around, where we had a lower resource language in our direction and it happened to be our target language.

2. Llama's output was only acceptable with the highest resource languages; otherwise, the outputs were completely disqualified from any use case, yet alone to be evaluated.

3. Some models, maybe not much efficient with some low-resource languages, including Persian; in the case of such languages, not only the amount of the shots did not improve the outputs' quality, but also any direction with that low resource language, fell behind directions with high resource languages on both sides. it must be because of the lack of training and exposure to those languages. Furthermore, lots of hallucination and content deviation, mixing languages, etc. were observed in that case. In Contrast, if our target language was a high resource language, especially English; the results with applying n-shot method most likely happened to be boosted.

4. Perplexity and after that GPT 4 were the best models in this setting. Claude showed the best capacity among all of them with applying a greater number of shots. Overall, in three directions, its efficiency increased but showed inconsistency in other ones.

5. Most of the transitions of changing error numbers occur either changing from zero shot to one shot or from two shot to few-shot.

## 3.4 Comparison of n-shot, prompt enhanced and their combinations

*3.4.1 Explanation.* In this section we have illustrated the tables below, to compare the effect of incorporating prompting frameworks in addition to the shots, with the original setting in which we solely used shots. the results for each of the language directions have been sorted, and for each one of them, we have described the results accordingly.

| Fa to Ru | | 0 Shot | 1 Shot | 2 Shot | Few Shot | |
|---|---|---|---|---|---|---|
| | GPT 3.5 | Decrease↓ (by 1) | Decrease↓ (by 1) | Decrease↓ (by 1) | | T1 |
| | | Decrease↓ (by 1) | Decrease↓ (by 1) | | | T2 |
| | | Decrease↓ (by 1) | Decrease↓ (by 1) | | | T3 |
| | | | | | Decrease↓ (by 1) | T4 |
| | | Decrease↓ (by 1) | | Decrease↓ (by 1) | Increase↑ (by 1) | T5 |
| | | | | | | T6 |
| | | Decrease↓ (by 1) | Decrease↓ (by 1) | Decrease↓ (by 1) | Decrease↓ (by 1) | T7 |
| | GPT 4 | | | | | T1 |
| | | | | | Decrease↓ (by 1) | T2 |
| | | | | | | T3 |
| | | Decrease↓ (by 1) | | | | T4 |
| | | | | | | T5 |
| | | | | | | T6 |
| | | | | | | T7 |
| | PaLM | | | | | T1 |
| | | | | | | T2 |
| | | | Increase↑ (by 3) | Increase↑ (by 3) | Increase↑ (by 3) | T3 |
| | | | | | | T4 |
| | | | | | | T5 |
| | | | | | | T6 |
| | | | | | | T7 |
| | Llama | Translation rejected not qualified for evaluation | | | | |
| | Claude | | Decrease↓ (by 1) | Decrease↓ (by 1) | Decrease↓ (by 1) | T1 |
| | | | | | | T2 |
| | | Increase↑ (by 3) | | | | T3 |
| | | Increase↑ (by 3) | Increase↑ (by 3) | Decrease↓ (by 1) | Increase↑ (by 2) | T4 |
| | | | | | | T5 |
| | | | | | | T6 |
| | | Decrease↓ (by 1) | Decrease↓ (by 1) | Decrease↓ (by 1) | Decrease↓ (by 1) | T7 |
| | Perplexity | Decrease↓ (by 1) | Increase↑ (by 1) | Decrease↓ (by 1) | Decrease↓ (by 1) | T1 |
| | | | | Increase↑ (by 1) | Increase↑ (by 1) | T2 |
| | | Increase↑ (by 1) | Increase↑ (by 2) | | | T3 |
| | | Increase↑ (by 1) | Decrease↓ (by 1) | | | T4 |
| | | Decrease↓ (by 1) | | Decrease↓ (by 1) | Decrease↓ (by 1) | T5 |
| | | Decrease↓ (by 1) | Decrease↓ (by 1) | Decrease↓ (by 1) | Decrease↓ (by 1) | T6 |
| | | Decrease↓ (by 2) | Decrease↓ (by 2) | Decrease↓ (by 2) | Decrease↓ (by 2) | T7 |

Results: A
As shown by the chart PaLM and GPT4 showed the least changes by adding our prompt frameworks.
GPT3.5, GPT4, and Perplexity efficiency improved.
Claude and PaLM efficiency decreased.

| Ru to Fa | | 0 Shot | 1 Shot | 2 Shot | Few Shot | |
|---|---|---|---|---|---|---|
| | GPT 3.5 | Decrease↓ (by 1) | Increase↑ (by 1) | Increase↑ (by 1) | Increase↑ (by 1) | T1 |
| | | | | | | T2 |
| | | | | | | T3 |
| | | Increase↑ (by 3) | Increase↑ (by 2) | Increase↑ (by 2) | | T4 |
| | | Increase↑ (by 3) | Increase↑ (by 4) | Increase↑ (by 4) | Increase↑ (by 4) | T5 |
| | | Decrease↓ (by 2) | Decrease↓ (by 2) | Decrease↓ (by 2) | Decrease↓ (by 3) | T6 |
| | GPT 4 | | | | Decrease↓ (by 1) | T1 |
| | | | | | | T2 |
| | | Decrease↓ (by 1) | Decrease↓ (by 1) | Decrease↓ (by 3) | Decrease↓ (by 3) | T3 |
| | | Increase↑ (by 1) | Increase↑ (by 1) | Increase↑ (by 6) | Increase↑ (by 2) | T4 |
| | | Decrease↓ (by 2) | Decrease↓ (by 2) | Decrease↓ (by 1) | Decrease↓ (by 1) | T5 |
| | | | | Decrease↓ (by 1) | | T6 |
| | PaLM | | | | | T1 |
| | | | | | Increase↑ (by 1) | T2 |
| | | | | | Decrease↓ (by 1) | T3 |
| | | | | | | T4 |
| | | | | | | T5 |
| | | Decrease↓ (by 1) | Decrease↓ (by 1) | | Decrease↓ (by 1) | T6 |
| | Llama | Translation rejected not qualified for evaluation | | | | |
| | Claude | | | | | T1 |
| | | | | | | T2 |
| | | Increase↑ (by 1) | | | | T3 |
| | | | | | | T4 |
| | | | | | | T5 |
| | | | | | | T6 |
| | Perplexity | | | | | T1 |
| | | | | | | T2 |
| | | Decrease↓ (by 1) | Decrease↓ (by 1) | Decrease↓ (by 1) | Decrease↓ (by 1) | T3 |
| | | | | | | T4 |
| | | | | | | T5 |
| | | Increase↑ (by 1) | Increase↑ (by 1) | Increase↑ (by 1) | Increase↑ (by 1) | T6 |

Results:
As shown by the chart, GPT4 and PaLM showed improvements.
Claude and Perplexity remained the same.
GPT 3.5 results were worse and contained more errors.

Fig. 31. Comparing methods, Fa to Ru    Fig. 32. Comparing methods, Ru to Fa

Fig. 33. Comparing methods, Eng to Fa.    Fig. 34. Comparing methods, Fa to Eng

Fig. 35. Comparing methods, Ru to Eng    Fig. 36. Comparing methods, Eng to Ru

*3.4.1 Results.* Overall Results of the Comparison are listed as below.

    1.    PaLM was the most efficient model, the efficiency improved with all our three stings; including increasing shots, applying a prompting framework, and a

combination of both, and this model could best analyze long texts, apply the stylistics nuances effectively; We hypothesize that this may be due to the fact that this model can further improve its bilingual abilities as its training has involved in multilingual data and can further improve this ability as it regulates and applies it more often; we also hypothesize that this capacity has improved analytical linguistics abilities as this model best handled the data sets which were given as examples and applied prompts regarding style and intonation more effectively compare to models with less multilingual abilities.

2. After that Perplexity AI was the best one to understand all our data, shots and comprehensive prompts. On the other hand, by adding more data its efficiency did not decrease, nor the model produced more errors; It was either improvements or minor changes.

3. When the target language was a low resource, and our model happened to be not good with that language; in our case here Claude, adding more data, shots and lengthy prompts, further decreased the efficiency and produced more errors.

4. GPT 3.5 showed sensitive behavior toward adding data to its prompts; explaining more complete, or adding shots most likely decreased its efficiency and generally it was not good at handling lengthy prompts and data.

5. Overall, a combination of both of improving prompt with a standard framework and adding shots alongside that framework, most likely happened to decrease the model's efficiency in translation; otherwise, the model has capacity on handling data with different patterns and ability to analyze lengthy texts.

## 3.4 Comparison of n-shot, prompt enhanced and their combinations based on Neural metrics

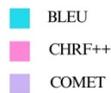

Fig. 37. Color Guides

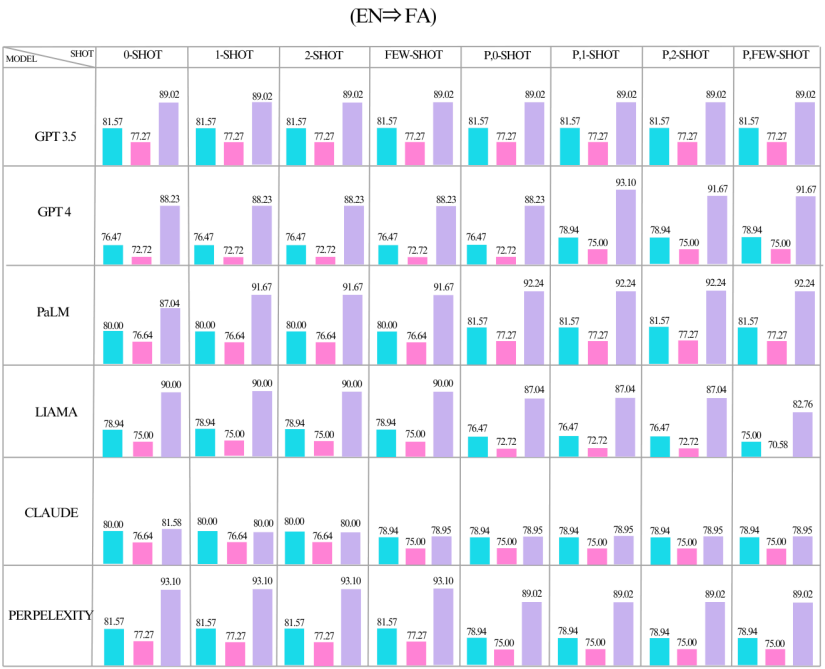

Fig. 38. Neural metric analysis on prompt enhanced n-shot method, English to Persian

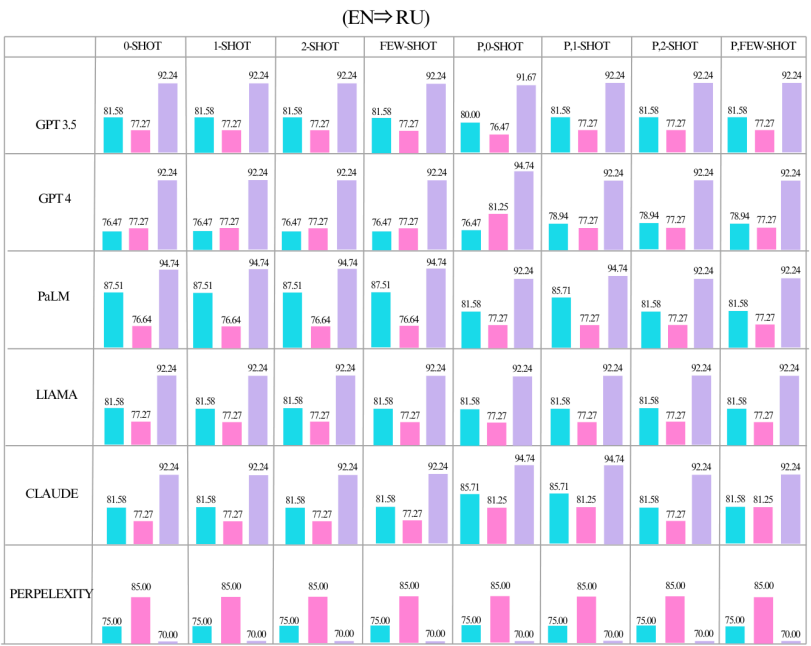

Fig. 39. Neural metric analysis on prompt enhanced n-shot method, English to Russian

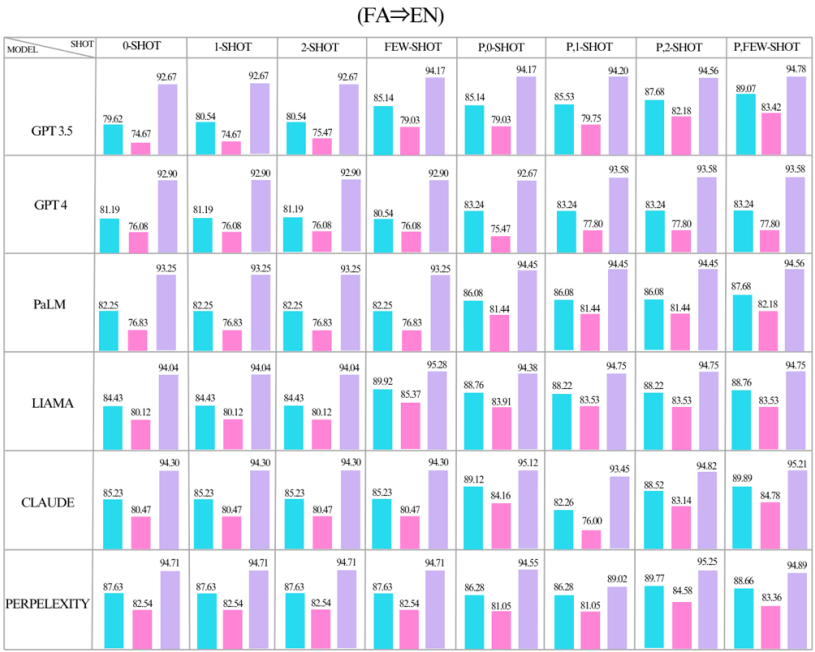

Fig. 40. Neural metric analysis on prompt enhanced n-shot method, Persian to English

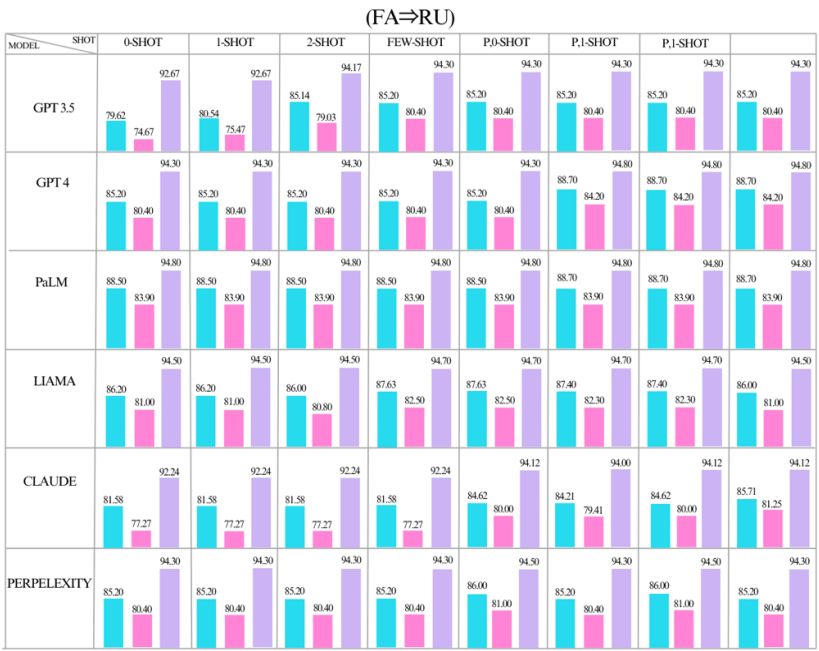

Fig. 41. Neural metric analysis on prompt enhanced n-shot method, Persian to Russian

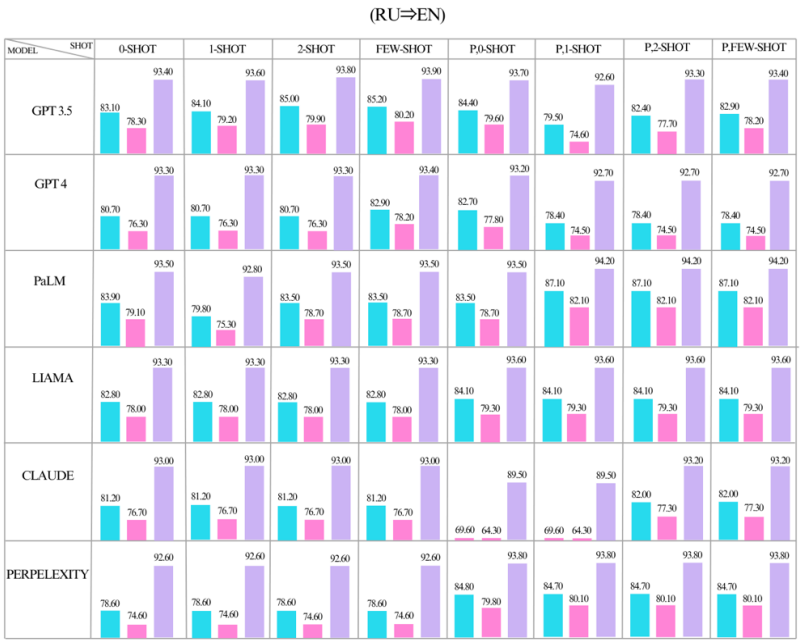

Fig. 42. Neural metric analysis on prompt enhanced n-shot method, Russian to English

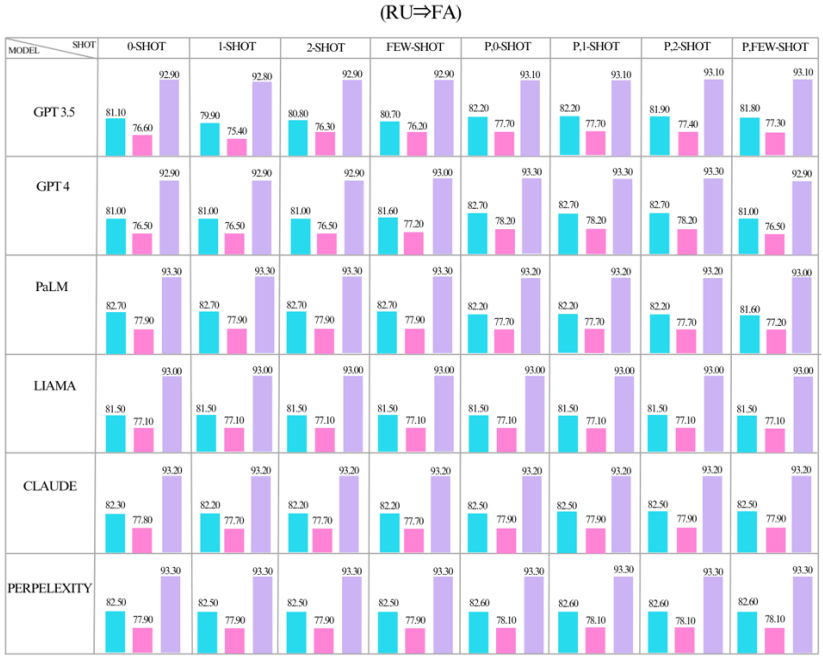

Fig. 43. Neural metric analysis on prompt enhanced n-shot method, Russian to Persian

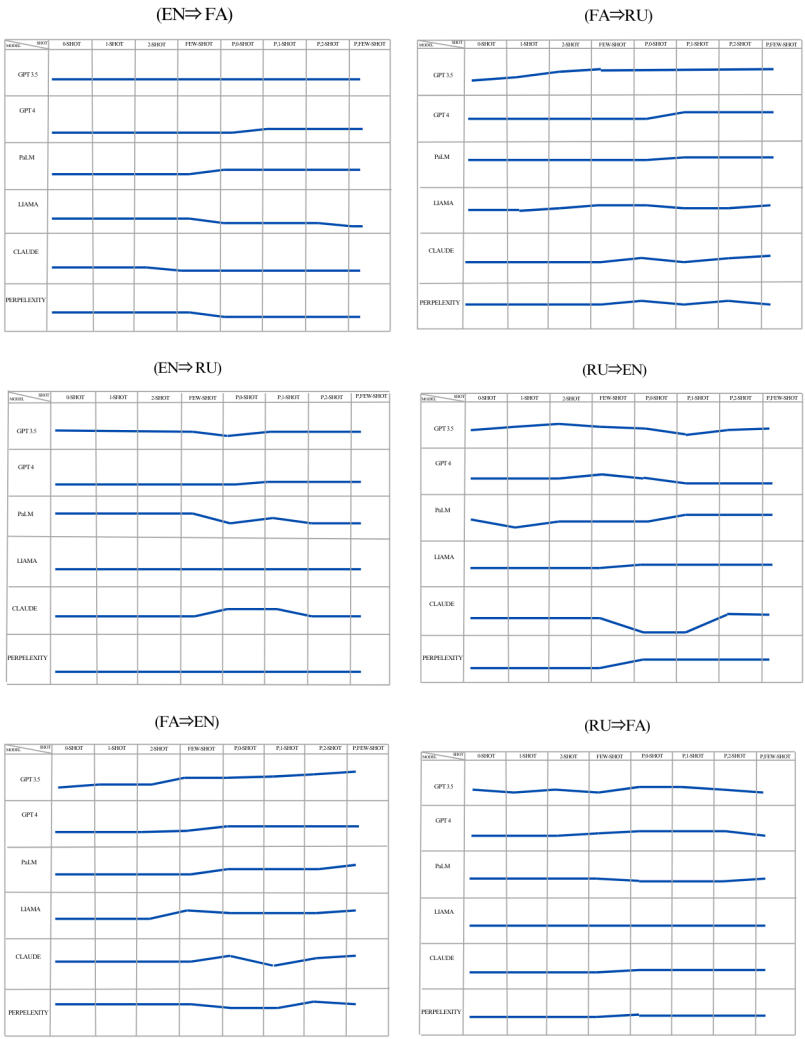

Fig. 44. Bleu Metric changes

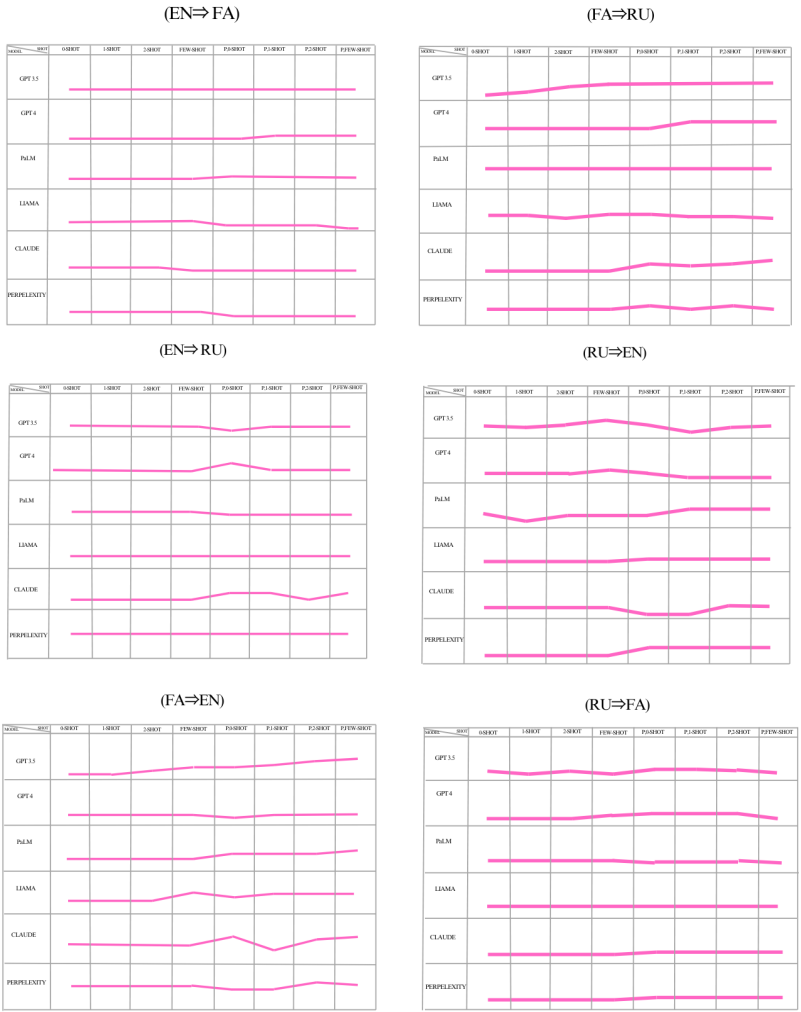

Fig. 45. Chrf++ Metric changes

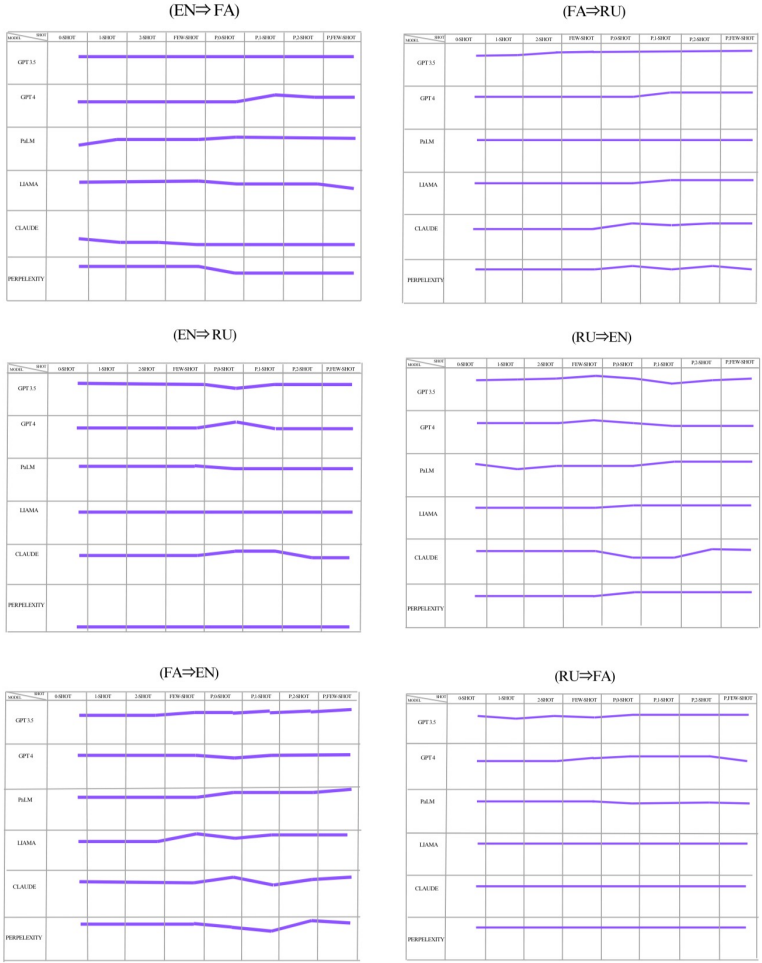

Fig. 46. COMET Metric changes

The Bleu score tended to increase with the models that perform better with analyzing long texts and their potential to develop multilingual capacity. In contrast, for models that operated weaker for more distant languages, the combination of adding a prompting framework, and in-context learning shots reduced their efficiency which aligns with our error analysis through human evaluation.

We also found that lexical comparison-based metrics such as BLEU or chrf++ inconsistently gave a few misleading signals, such limitations stem from the inherent challenges of evaluating natural language generation systems, especially for complex tasks such as MT; therefore, we complemented our automatic evaluation with a comprehensive analysis, considering all metrics together, as well as human evaluation as our main methodology, and qualitative analysis to cover a broad range of phenomena.

In general, human error analysis, and neural network metrics are not enough for evaluation, when it comes to identifying a translation output that is useful for a translator to be used in their final work. We observed the fluency and the efficiency of the overall text improved effectively when a tailored, and standard prompting framework is used, but this is totally different from being flawless, and just as we proposed above, besides their fluency and the amazing guidance they offer to the translator, this does not mean they do not contain errors; hence, the translator needs to put more careful attention into the post-editing process of NMT raw outputs; and somehow identifying errors and the post-editing process becomes much more tricky when the overall text seem to have accuracy and fluency; therefore we proposed our analogy for our selected language combinations, hopefully to help translators adapt better to post-editing brand-new LLMs raw outputs.

## 4   CONCLUSIONS

In this work, we presented a study of the machine translation capabilities of the trending and popular LLMs of 2023. Our investigation covered 6 language directions, which are all combinations of English, Russian, and Persian, across a number of selected shots for the purpose of evaluating in-context learning with our selected LLMs; all of which facilitated a broad understanding of the models' general performance for our desired directions. Furthermore, we used a standard translation-tailored prompting framework to compare the results on the performance of the LLMs between these two settings; in the same vein, we also presented the results for our third setting which contained a combination of the two other methods and presented the results for all of these settings respectively as well as a whole together. To provide a thorough evaluation of the models, we employed both human evaluations and the latest neural network-based automatic evaluation metrics. In addition, we conducted extensive analysis, providing an in- depth comparison between all of these three methods both in sense of the error rate with human analysis and their comparisons to state-of-the-art neural metrics.

Here we summarize some interesting findings and useful recipes on popular LLMs as MT tools:

• The efficiency of the model not just was relevant and dependent to the language direction but also dependent on the target language; if the target language was a high-resource language, for instance English, the efficiency improved significantly.

• N-shot method improved the efficiency of a model like GPT 3.5, but this method not necessarily improved efficiency with of all other models in all of the directions we experimented with.

• We found that prompting templates and demonstration example selection both have substantial impact on outputs. Prompting examples correlated significantly with the performance of our model; for example, careful selection benefited translation to some extent but not consistently.

• By evaluating the translation among three selected languages on our test sets, we found that PaLM2 performed competitively on both high-resource language like English or Russian as well as more under-resourced language like Persian. GPT4 despite amazing results on high resource languages, lagged with a more distant language, and for language pairs that were both low-resource and from different families [58].

Our findings align with Previous researchers [23], [21], [39] who have found that an LLM like ChatGPT performs competitively with commercial translation products (e.g., Google Translate and Microsoft Translator) on high-resource languages, but proved to have less capabilities for low-resource and distant languages. Furthermore, PaLM was the best model to perform translation prompting frameworks and analyze the examples given via in-context learning in an n-shot setting. We hypothesize that this is due to the large multilingual data proportion in its training data, which is 78% English and 22% for other languages [48], while GPT data proportion is only 7% non-English [4].

• the efficiency of PaLM improved with all our three stings; including increasing shots, applying a prompt framework, and a combination of both, and this model can best analyze long texts, apply the stylistics nuances obvious and clear as well; We hypothesize that this may be due to the fact that this model can further improve its bilingual abilities as its training has involved in multilingual data and can further improve this ability as it regulates and apply it more often; we also hypothesize that this capacity has improved analytical linguistics abilities as this model best handled the data sets which were given as examples and applied prompts regarding style and intonation more effectively compare to models with less multilingual abilities.

• after that Perplexity AI was the best one to understand prompts completely, including shots and explanations. On the other hand, by adding more data its efficiency did not decrease, nor the model produced more errors; it was either improvements or minor changes.

• When the target language was a low resource one, and our model happened to have little data with that; adding more data, shots and lengthy prompts, further decreased efficiency and produced more errors. Furthermore, lots of hallucination and content deviation, mixing languages, etc. were observed in the case of this scenario. In Contrast, if the target language was a high resource language, especially English; the results with applying tailored prompts like n-shot method and using a prompting framework, most likely happened to be boosted.

• Some of the models like Claude, maybe not much efficient with some low-resource languages, including Persian; any language direction with low resource languages, fell behind directions with high resource languages on both sides. it must be because of the lack of training and exposure to those languages, so the hallucination and error rate most likely to be abundant in those scenarios.

• most of the transitions regarding error numbers, either increasing or decreasing, occurred either changing from zero shot to one shot or from two shot to few-shot.

• GPT 3.5 showed highly sensitive behavior regarding data addition to its prompts; explaining too much, or adding shots most likely happened to decrease its efficiency and generally it was not good at handling lengthy prompts and data.

• Overall, a combination of both using a prompting frame work with adding in-context learning shots, most likely happened to decrease the model's efficiency in translation; except that the model is capable of handling data with different patterns and ability to analyze lengthy texts.

Overall, our study provides valuable insights into the strengths and weaknesses of trending LLMs of the time this paper is being written for machine translation. This work opens up opportunities for future improvements and developments in this field. We investigated how these models can transform the world of machine translation tasks along with multiple strengths it applies as a result of generative content. We demonstrated that many of these models excel at translating well-represented languages in their training data, but they face challenges with less-resourced languages, except they have been trained on a reasonable amount of multilingual/ bilingual data.

Furthermore, the interpretation of the results obtained allowed us to identify recurring patterns of errors, thus providing an evaluation of the raw output.

What emerges from this study is that, although we observed critical errors depending on the nature of the models, that we tested respectively; with some instances the raw translation turned out to be satisfactory without having to apply post editing, on the other hand some turned out to be far from the bottom line of having the quality for being evaluated, and in no case, we could adopt them nor get help from them. The evaluation conducted for this research provides translation professionals and scholars with an insight of the performance of LLMs as MT tools, through a list of predominant errors, which correspond to aspects that should be carefully controlled at the post-editing stage in the English-Russian-Persian combination directions.

In case of using a model, that has been trained on large corpora of multi lingual data for the purpose of using it on a direction of high/acceptable resource language they most likely have adopted human level translation, but it is always advised to use post-editing stage with an efficient translator, and we are far from a scenario in which we can overlook or skip this stage for purposes that need a proper well drafted text. But even in the mentioned scenario, human translators will play a key role, as the development of more efficient MT tools will mostly depend on collaboration between computer engineers and professional translators. Therefore, it seems essential to implement an 'orchestrated symbiosis' in the words of Bawa- Mason [3]; it is crucial that translators do not consider technology as a competitor but as a means to enhance their performance. Working hand in hand with computer engineers is essential to improve LLM-based MT systems. Such collaboration would allow engineers to understand better the equivalence issues between languages as well as typical translation problems, and thus to design new systems able to provide even better results.

The analysis conducted for this project provides a list of features that NMT specialists should endeavor to improve when developing new tools (language in context, the importance of specialized terminology, etc.). Furthermore, receiving feedback from linguists working with NMT systems is also essential for the implementation of more sophisticated automatic metrics suitable for the evaluation of more recent MT tools. As a future research direction, we propose to tackle the capacity of LLMs for post editing and using them to provide specific purpose-based analogies.

**5         Limitations:**

We admit that this study is far from complete from various aspects; in order to make it more reliable:

• Coverage of Test Data: Currently, we randomly selected our samples from each test set for evaluation due to the lack of reliable translation equivalences that has been long adopted the culture difference and language nuances. Therefore, we still report the current results with references we found with the highest level of equivalence. Furthermore, the size of our data set is small, because we aimed to make the human evaluation process consistent, and we did not want to make the process tedious; If we could access more multilingual data, linguists and translators with similar backgrounds in our language direction, who could collaborate on designing an error analogy pattern, the results could have been even more in detail. It would also be relevant to extend this study to different text genres in order to verify whether it would show similar results.

• The same goes for the linguistic combination. This research project only focused on the English-Russian-Persian language pairs; It would therefore be relevant to evaluate the same experiment for more distant languages and more diverse combinations. This could help to identify the strengths and weaknesses of our models, for other language pairs, in addition to the ones that were studied. The conclusion of this study should be taken in this context and not generalized to other languages without further evaluation.

• We conducted our evaluation using only reliable test sets and baselines. While a more comprehensive evaluation is needed to be taken on our directions, we should be cautious about drawing conclusions from low quality test sets or baselines which are usually dominating the research results for low resource languages.

• By querying and prompting our selected LLMs multiple times, we found that the results of the same query may vary across multiple trials, which brings randomness to the evaluation results. For better reliable results, it is best to repeat the translation multiple times for each test set and report the average result.

• Evaluation Metrics: The results here are calculated by automatic metrics with single references, which may not convey some characteristics of translation properly. Therefore, our human evaluation provides more insights for our desired comparison.

• We established a typology as our methodology for human evaluation, which enables us to analyze linguistics and translation aspects as far as needed; Beyond this point, establishing a new error typology makes the experiment hardly reproducible and incomparable to other previously experimented typologies. Therefore, it is a good idea to consider previously tested typologies, directions, and experimental settings. We recommend that readers consider the overall evaluations as a whole, rather than relying solely on a specific metric, to better understand the quality of LLMs' MT capabilities.

In future work, we would like to experiment with more diverse prompting types and techniques to further improve the performance of LLMs in MT, and conduct more in-depth comparisons and discussions.

## REFERENCES

[1] S. Agrawal, C. Zhou, M. Lewis, L. Zettlemoyer, and M. Ghazvininejad. 2022. In-context Examples Selection for Machine Translation. In Findings of the Association for Computational Linguistics: ACL 2023, pages 8857–8873, Toronto, Canada. Association for Computational Linguistics. https://doi.org/10.18653/v1/2023.findings-acl.564

[2] F. Akhbardeh, A. Arkhangorodsky, M. Biesialska, and others. 2021. Findings of the 2021 conference on machine translation (WMT21). In Proceedings of the Sixth Conference on Machine Translation, pages 1–88. Online. Association for Computational Linguistics. https://aclanthology.org/2021.wmt-1.1

[3] S. Bawa-Mason, L. Bywood, C. Gittins, and others. 2018. Translators in the Digital Era: What Kind of Jobs Will We Have Ten Years from Now? Presented at The Language Show, Olympia, London, UK. https://aclanthology.org/W19-87.pdf

[4] T. B. Brown, B. Mann, N. Ryder, and others. 2020. "Language models are few-shot learners." Advances in neural information processing systems. https://proceedings.neurips.cc/paper_files/paper/2020/file/1457c0d6bfcb4967418bfb8ac142f64a-Paper.pdf

[5] C. Callison-Burch, C. Fordyce, P. Koehn, C. and others. 2007. (Meta-) Evaluation of Machine Translation. In Proceedings of the Second Workshop on Statistical Machine Translation, Prague, Czech Republic. Association for Computational Linguistics. https://aclanthology.org/W07-0718

[6] C. Callison-Burch, M. Osborne, and P. Koehn. 2006. "Re-Evaluating the Role of BLEU in Machine Translation Research." In Proceedings of the 11th Conference of the European Chapter of the Association for Computational Linguistics, ACL, Trento, Italy. Association for Computational Linguistics. https://aclanthology.org/E06-1032

[7] A. Chowdhery, S. Narang, J. Devlin, and others. 2022. "Palm: Scaling language modeling with pathways." Journal of Machine Learning Research. https://www.jmlr.org/papers/volume24/22-1144/22-1144.pdf

[8] Marta R. Costa-jussà, J. Cross, O. Çelebi, and others. 2022. "No language left behind: Scaling human-centered machine translation." arXiv preprint arXiv:2207.04672. https://doi.org/10.48550/arXiv.2207.04672

[9] J. Daems, S. Vandepitte, R. J. Hartsuiker, and L. Macken. 2017. "Identifying the Machine Translation Error Types with the Greatest Impact on Post-Editing Effort." Frontiers in Psychology, vol. 8, 1282. http://dx.doi.org/10.3389/fpsyg.2017.01282

[10] G. Doddington. 2002. "Automatic Evaluation of Machine Translation Quality Using n-gram Co-occurrence Statistics." In Proceedings of the Second Conference on Human Language Technology Research (HLT '02). ACL, San Francisco, CA, USA, 138–145. http://dx.doi.org/10.3115/1289189.1289273

[11] Q. Dong, L. Li, D. Dai, and others. 2022. "A Survey on In-context Learning." arXiv [cs.CL]. https://arxiv.org/abs/2301.00234. https://doi.org/10.48550/arXiv.2301.00234

[12] Marie Escribe. 2019. Human Evaluation of Neural Machine Translation: The Case of Deep Learning. In Proceedings of the Human-Informed Translation and Interpreting Technology Workshop (HiT-IT 2019), pages 36–46, Varna, Bulgaria. Incoma Ltd., Shoumen, Bulgaria. https://aclanthology.org/W19-8705. https://doi.org/10.26615/issn.2683-0078.2019_005

[13] A. Fan, S. Bhosale, H. Schwenk, and others. 2021. "Beyond English-centric multilingual machine translation." The Journal of Machine Learning Research. https://www.jmlr.org/papers/volume22/20-1307/20-1307.pdf. https://doi.org/10.48550/arXiv.2010.11125

[14] A. Font Llitjós, Jaime G. Carbonell, and A. Lavie. 2005. A framework for interactive and automatic refinement of transfer-based machine translation. In Proceedings of the 10th EAMT Conference: Practical applications of machine translation, Budapest, Hungary. European Association for Machine Translation. https://aclanthology.org/2005.eamt-1.13

[15] M. Freitag, R. Rei, N. Mathur, and others. 2022. "Results of WMT22 metrics shared task: Stop using BLEU – neural metrics are better and more robust." In Proceedings of the Seventh Conference on Machine Translation (WMT), pages 46–68, Abu Dhabi, United Arab Emirates (Hybrid). Association for Computational Linguistics. https://aclanthology.org/2022.wmt-1.2

[16] M. Freitag, G. Foster, D. Grangier, and others. 2021. "Experts, errors, and context: A large-scale study of human evaluation for machine translation." Transactions of the Association for Computational Linguistics. https://doi.org/10.1162/tacl_a_00437. https://doi.org/10.48550/arXiv.2104.14478.

[17] X. Garcia, Y. Bansal, C. Cherry, and others. 2023. "The unreasonable effectiveness of few-shot learning for machine translation." In Proceedings of the 40th International Conference on Machine Learning (ICML'23), Vol. 202. JMLR.org, Article 438, 10867–10878. https://proceedings.mlr.press/v202/garcia23a/garcia23a.pdf

[18] T. Goyal, J. J. Li, and G. Durrett. 2022. "News summarization and evaluation in the era of GPT-3." arXiv preprint arXiv:2209.12356. https://doi.org/10.48550/arXiv.2209.12356

[19] Francisco J. Guzman, A. Abdelali, I. Temnikova, and others. 2015. How do Humans Evaluate Machine Translation. In Proceedings of the Tenth Workshop on Statistical Machine Translation, pages 457–466, Lisbon, Portugal. Association for Computational Linguistics. http://dx.doi.org/10.18653/v1/W15-3059

[20] H. Hassan, A. Aue, C. Chen, and others. 2018. "Achieving Human Parity on Automatic Chinese to English News Translation." Microsoft AI & Research. arXiv:1803.05567. https://doi.org/10.48550/arXiv.1803.05567.

[21] A. Hendy, M. Abdelrehim, A. Sharaf, and others. 2023. "How good are GPT models at machine translation? A comprehensive evaluation." https://arxiv.org/abs/2302.09210. https://doi.org/10.48550/arXiv.2302.09210

[22] Z. He, X. Wang, Z. Tu, and others. 2022. "Tencent AI Lab-Shanghai Jiao Tong University low-resource translation system for the WMT22 translation task." In Proceedings of the Seventh Conference on Machine Translation (WMT), pages 260–267, Abu Dhabi, United Arab Emirates (Hybrid). Association for Computational Linguistics. https://aclanthology.org/2022.wmt-1.18

[23] W. Jiao, W. Wang, J.-t. and others. 2023. "Is ChatGPT a good translator? A preliminary study." arXiv preprint arXiv:2301.08745. https://doi.org/10.48550/arXiv.2301.08745

[24] W. Jiao, Z. Tu, J. Li, and others. 2022a. Tencent's Multilingual Machine Translation System for WMT22 Large-Scale African Languages. In Proceedings of the Seventh Conference on Machine Translation (WMT), pages 1049–1056, Abu Dhabi, United Arab Emirates (Hybrid). Association for Computational Linguistics. https://aclanthology.org/2022.wmt-1.102

[25] W. Jiao, X. Wang, S. He, Z. Tu, I. King and M. R. Lyu, "Exploiting Inactive Examples for Natural Language Generation with Data Rejuvenation," in IEEE/ACM Transactions on Audio, Speech, and Language Processing, vol. 30, pp. 931-943, 2022, doi: 10.1109/TASLP.2022.3153269.

[26] W. Jiao, X. Wang, Z. Tu, and others. 2021. Self-Training Sampling with Monolingual Data Uncertainty for Neural Machine Translation. In Proceedings of the 59th Annual Meeting of the Association for Computational Linguistics and the 11th International Joint Conference on Natural Language Processing (Volume 1: Long Papers), pages 2840–2850, Online. Association for Computational Linguistics.

[27] M. Johnson, M. Schuster, Q. Le, and others. 2017. Google's Multilingual Neural Machine Translation System: Enabling Zero-Shot Translation. Transactions of the Association for Computational Linguistics, 5:339–351. https://doi.org/10.1162/tacl_a_00065

[28] J. Kaplan, S. McCandlish, T. Henighan, T. and others. 2020. "Scaling laws for neural language models." arXiv preprint arXiv:2001.08361. https://doi.org/10.48550/arXiv.2001.08361

[29] Y. J. Kim, A. A. Awan, A. Muzio, and others. 2021. "Scalable and efficient moe training for multitask multilingual models." arXiv preprint arXiv:2109.10465. https://doi.org/10.48550/arXiv.2109.10465

[30] T. Kocmi and C. Federmann. 2023. Large language models are state-of-the-art evaluators of translation quality. In Proceedings of the 24th Annual Conference of the European Association for Machine Translation, pages 193–203, Tampere, Finland. European Association for Machine Translation. https://aclanthology.org/2023.eamt-1.19

[31] P. Koehn. 2009. Human translation and machine translation. In Proceedings of the 6th International Workshop on Spoken Language Translation: Plenaries, Tokyo, Japan. https://aclanthology.org/2009.iwslt-keynotes.1

[32] Q. Lu, B. Qiu, L. Ding, and others. 2023. "Error analysis prompting enables human-like translation evaluation in large language models: A case study on ChatGPT." https://arxiv.org/abs/2303.13809. https://doi.org/10.48550/arXiv.2303.13809

[33] Y. Moslem, R. Haque, J. Kelleher, and A. Way. 2022. "Domain-Specific Text Generation for Machine Translation." In Proceedings of the 15th biennial conference of the Association for Machine Translation in the Americas (Volume 1: Research Track), pages 14–30, Orlando, USA, Association for Machine Translation in the Americas. https://aclanthology.org/2022.amta-research.2

[34] Yasmin Moslem, Rejwanul Haque, John D. Kelleher, and Andy Way. 2023. Adaptive Machine Translation with Large Language Models. In Proceedings of the 24th Annual Conference of the European Association for Machine Translation, pages 227–237, Tampere, Finland. European Association for Machine Translation. https://aclanthology.org/2023.eamt-1.22

[35] S. Nirenburg; Harold L. Somers; and others. 2003. "A Framework of a Mechanical Translation between Japanese and English by Analogy Principle," in Readings in Machine Translation, MIT Press. https://doi.org/10.7551/mitpress/5779.001.0001

[36] L. Ouyang, J. Wu, X. Jiang, D. and others. 2022. "Training language models to follow instructions with human feedback." Advances in Neural Information Processing Systems. https://proceedings.neurips.cc/paper_files/paper/2022/file/b1efde53be364a73914f58805a001731-Paper-Conference.pdf

[37] Hazel M. Pan. 2016. How BLEU Measures Translation and Why It Matters. Slator.

[38] K. Papineni, S. Roukos, T. Ward, and W.-J. Zhu. 2002. "Bleu: a method for automatic evaluation of machine translation." In Proceedings of the 40th Annual Meeting of the Association for Computational Linguistics, pages 311–318, Philadelphia, Pennsylvania, USA. Association for Computational Linguistics. https://doi.org/10.3115/1073083.1073135

[39] K. Peng, L. Ding, Q. Zhong, L. and others. 2023. "Towards making the most of ChatGPT for machine translation." arXiv preprint arXiv:2303.13780. https://doi.org/10.48550/arXiv.2303.13780

[40] M. Popović. 2017. chrF++: words helping character n-grams. In Proceedings of the Second Conference on Machine Translation, pages 612–618, Copenhagen, Denmark. Association for Computational Linguistics. https://doi.org/10.18653/v1/W17-4770

[41] C. Qin, A. Zhang, Z. Zhang, and others. 2023. "Is ChatGPT a general-purpose natural language processing task solver?" In Proceedings of the 2023 Conference on Empirical Methods in Natural Language Processing, pages 1339–1384, Singapore. Association for Computational Linguistics. https://doi.org/10.18653/v1/2023.emnlp-main.85

[42] A. Radford, J. Wu, R. Child, and others. 2019. "Language models are unsupervised multitask learners." Technical report, OpenAi. https://insightcivic.s3.us-east-1.amazonaws.com/language-models.pdf

[43] V. Raunak, A. Menezes, and M. Junczys-Dowmunt. 2021. "The curious case of hallucinations in neural machine translation." In Proceedings of the 2021 Conference of the North American Chapter of the Association for Computational Linguistics: Human Language Technologies, pages 1172–1183, Online. Association for Computational Linguistics. 10.18653/v1/2021.naacl-main.92

[44] R. Rei, C. Stewart, A. C. Farinha, and A. Lavie. 2020. "COMET: A neural framework for MT evaluation." In Proceedings of the 2020 Conference on Empirical Methods in Natural Language Processing (EMNLP), pages 2685–2702, Online. Association for Computational Linguistics. https://doi.org/10.18653/v1/2020.emnlp-main.213

[45] G. Sanders, M. Przybocki, N. Madnani, and M. Snover. 2011. Human Subjective Judgments. Handbook of Natural Language Processing and Machine Translation, pages 750–759. Springer.

[46] R. Sennrich. 2016. "Neural Machine Translation: Breaking the Performance Plateau." Presented at the META-FORUM 2016, Lisbon, Portugal.

[47] A. Schioppa, D. Vilar, A. Sokolov, and K. Filippova. 2021. "Controlling machine translation for multiple attributes with additive interventions." In Proceedings of the 2021 Conference on Empirical Methods in Natural Language Processing, pages 6676–6696, Online and Punta Cana, Dominican Republic. Association for Computational Linguistics. https://doi.org/10.18653/v1/2021.emnlp-main.535

[48] F. Shi, M. Suzgun, M. Freitag, and others. 2022. "Language models are multilingual chain-of-thought reasoners." https://arxiv.org/abs/2210.03057. https://doi.org/10.48550/arXiv.2210.03057

[49] M. Snover, B. Dorr, R. Schwartz, and others. 2006. A Study of Translation Edit Rate with Targeted Human Annotation. In Proceedings of the 7th Conference of the Association for Machine Translation in the Americas: Technical Papers, pages 223–231, Cambridge, Massachusetts, USA. Association for Machine Translation in the Americas. https://aclanthology.org/2006.amta-papers.25

[50] S. Stymne, H. Danielsson, S. Bremin, and others. 2012. "Eye Tracking as a Tool for Machine Translation Error Analysis." In Proceedings of the International Conference on Language Resources and Evaluation, Istanbul, Turkey. European Language Resources Association (ELRA). http://www.lrec-conf.org/proceedings/lrec2012/pdf/192_Paper.pdf

[51] Teven L. Scao, A. Fan, C. Akiki, and others. 2022. BLOOM: A 176B-Parameter Open-Access Multilingual Language Model. arXiv [cs.CL], https://doi.org/10.48550/arXiv.2211.05100.

[52] H. Touvron, T. Lavril, G. Izacard, and others. 2023. "LLaMA: Open and Efficient Foundation Language Models." https://arxiv.org/abs/2302.13971. https://doi.org/10.48550/arXiv.2302.13971

[53] J. Turian, L. Shen, and I. D. Melamed. 2003. "Evaluation of Machine Translation and its Evaluation." In Proceedings of Machine Translation Summit IX, New Orleans, LA, USA. https://aclanthology.org/2003.mtsummit-papers.51

[54] D. Vilar, M. Freitag, C. Cherry, and others. 2022. "Prompting palm for translation: Assessing strategies and performance In Proceedings of the 61st Annual Meeting of the Association for Computational Linguistics

(Volume 1: Long Papers), pages 15406–15427, Toronto, Canada. Association for Computational Linguistics. https://doi.org/10.18653/v1/2023.acl-long.859

[55] D. Vilar, G. Leusch, H. Ney, and R. E. Banchs. 2007. "Human Evaluation of Machine Translation Through Binary System Comparisons." In Proceedings of the Second Workshop on Statistical Machine Translation, Prague, Czech Republic. Association for Computational Linguistics. https://aclanthology.org/W07-0713

[56] D. Vilar, J. Xu, L. Fernando D'Haro, and Hermann Ney. 2006. Error Analysis of Statistical Machine Translation Output. In Proceedings of the Fifth International Conference on Language Resources and Evaluation (LREC'06), Genoa, Italy. European Language Resources Association (ELRA). http://www.lrec-conf.org/proceedings/lrec2006/pdf/413_pdf.pdf.

[57] J. Wang, Y. Liang, F. Meng, and others. 2023. Cross-lingual summarization via chatgpt. http://dx.doi.org/10.13140/RG.2.2.23574.83527

[58] W. Wang, W. Jiao, S. Wang, Z. Tu, and M. R. Lyu. 2022. "Understanding and mitigating the uncertainty in zero-shot translation." https://arxiv.org/abs/2205.10068. https://doi.org/10.48550/arXiv.2205.10068

[59] S. Wang, Z. Tu, Z. Tan, and others. 2021. "Language Models are Good Translators." arXiv preprint. https://arxiv.org/abs/2106.13627. https://doi.org/10.48550/arXiv.2106.13627

[60] J. Wei, X. Wang, D. Schuurmans, and others. 2022. Chain of thought prompting elicits reasoning in large language models. NeurIPS 2022.

[61] J. White, T. O'Connell, and F. O'Mara. 1994. "The ARPA MT Evaluation Methodologies: Evolution, Lessons, and Future Approaches." In Proceedings of the Association for Machine Translation in the Americas Conference, Columbia, Maryland, USA. https://aclanthology.org/1994.amta-1.25

[62] C. Zan, K. Peng, L. Ding, B. Qiu. 2022. "Vega-MT: The JD explore academy machine translation system for WMT22." In Proceedings of the Seventh Conference on Machine Translation (WMT), pages 411–422, Abu Dhabi, United Arab Emirates (Hybrid). Association for Computational Linguistics. https://aclanthology.org/2022.wmt-1.37

[63] H. Zhang, R. Cao, L. Chen, and others. 2023. ACT-SQL: In-Context Learning for Text-to-SQL with Automatically-Generated Chain-of-Thought. In Findings of the Association for Computational Linguistics: EMNLP 2023, pages 3501–3532, Singapore. Association for Computational Linguistics. https://doi.org/10.18653/v1/2023.findings-emnlp.227

[64] T. Zhang, V. Kishore, F. Wu, and others 2020. "Bertscore: Evaluating text generation with bert." In ICLR. https://arxiv.org/abs/1904.09675. https://doi.org/10.48550/arXiv.1904.09675

[65] J. Zhang and C. Zong. 2016. "Exploiting source-side monolingual data in neural machine translation." In Proceedings of the 2016 Conference on Empirical Methods in Natural Language Processing. pages 1535–1545, Austin, Texas. Association for Computational Linguistics. https://doi.org/10.18653/v1/D16-1160

[66] Q. Zhong, L. Ding, J. Liu, and others. 2023. Can chatgpt understand too? a comparative study on chatgpt and fine-tuned bert. arXiv preprint. https://doi.org/10.48550/arXiv.2302.10198